\documentclass{article}

\usepackage[main, final]{neurips_2026}

\usepackage[utf8]{inputenc}
\usepackage[T1]{fontenc}
\usepackage{hyperref}
\usepackage{url}
\usepackage{booktabs}
\usepackage{amsfonts}
\usepackage{nicefrac}
\usepackage{microtype}
\usepackage{xcolor}
\usepackage{graphicx}
\usepackage{subcaption}
\usepackage{amsmath}
\usepackage{amssymb}
\usepackage{mathtools}
\usepackage{amsthm}
\usepackage{etoc}
\usepackage{tikz}
\usetikzlibrary{positioning,fit,calc}

\usepackage[capitalize,noabbrev]{cleveref}

\theoremstyle{plain}
\newtheorem{theorem}{Theorem}[section]
\newtheorem{proposition}[theorem]{Proposition}
\newtheorem{lemma}[theorem]{Lemma}
\newtheorem{corollary}[theorem]{Corollary}
\theoremstyle{definition}
\newtheorem{definition}[theorem]{Definition}

\theoremstyle{remark}
\newtheorem{remark}[theorem]{Remark}

\usepackage[disable,textsize=tiny]{todonotes}
\newcommand{\debug}[1]{#1}		

\newcommand{\newmacro}[2]{\newcommand{#1}{{\debug{#2}}}}		

\newmacro{\Ocal}{\mathcal{O}}
\newmacro{\Scal}{\mathcal{S}}
\newmacro{\Lcal}{\mathcal{L}}
\newmacro{\Ecal}{\mathcal{E}}
\newmacro{\Ccal}{\mathcal{C}}
\newmacro{\Dcal}{\mathcal{D}}
\newmacro{\Rcal}{\mathcal{R}}
\newmacro{\trace}{\text{trace}}
\newmacro{\cx}{\text{cx}}
\newmacro{\FAQ}{\textbf{FAQ}}
\newmacro{\Proj}{\textbf{Proj}}
\newmacro{\GAP}{\text{GAP}}
\newmacro{\GM}{\text{GM}}
\newmacro{\GW}{\text{GW}}
\newmacro{\GH}{\text{GH}}
\newmacro{\GHS}{\text{GHS}}
\newmacro{\R}{\mathbb{R}}
\newmacro{\E}{\mathbb{E}}
\newmacro{\bone}{\mathbf{1}}
\newmacro{\noise}{\text{noise}}
\newmacro{\score}{\text{score}}
\newmacro{\algo}{\text{algo}}
\newmacro{\acc}{\textbf{acc}}
\newmacro{\nce}{\textbf{nce}}
\newmacro{\Bernoulli}{\text{Bernoulli}}
\newmacro{\Pb}{\mathbb{P}}

\title{Chaining 2-FWL GNNs for Combinatorial Graph Alignment}

\author{%
  Marc Lelarge\\
  INRIA, \'Ecole Normale Sup\'erieure, PSL Research University\\
  Paris, France\\
  \texttt{marc.lelarge@inria.fr}
}

\begin{document}

\maketitle

\begin{abstract}
For the combinatorial graph alignment problem (GAP) --- finding the node correspondence that maximizes the number of common edges ($\nce$) between two unlabeled graphs --- the classical Frank--Wolfe heuristic $\FAQ$ remains a strong baseline once it is properly initialized, while existing GNN approaches struggle in the purely structural setting. We introduce a chaining procedure: a sequence of Folklore-type (2-FWL) GNNs in which each network is trained with cross-entropy after decoding the previous network's similarity matrix and ranking nodes by their current alignment quality. This non-differentiable ranking step injects discrete combinatorial feedback at every stage; at inference, we iterate the final network and keep the candidate with highest observed $\nce$. On sparse Erd\H{o}s--R\'enyi graphs at noise level $0.25$, chained FGNNs used to initialize $\FAQ$ reach $85\%$ accuracy versus $13\%$ for $\FAQ$ initialized from the convex relaxation, and essentially $0\%$ for prior GNN methods. On correlated regular graphs, where MPNNs with constant features produce identical node embeddings (1-WL fails to refine) and $\FAQ$'s convex initialization is degenerate, chaining is the only method among those we evaluate, including Gromov--Wasserstein solvers, that recovers a non-trivial alignment.  On three real-world benchmarks (yeast PPI, coauthorship, and road networks), recent comparisons underestimate $\FAQ$ by initializing it from a uniform doubly stochastic matrix; initialized from the convex relaxation it already surpasses prior reported numbers, and dataset-specific chained FGNNs further improve on this strengthened baseline. Code and pretrained chains available at: \url{https://github.com/mlelarge/chaining-gnn-graph-alignment}.
\end{abstract}

\section{Introduction}

Graph neural networks are attractive for graph combinatorial problems because they can learn structural representations from simulated instances, but in graph alignment they have not yet matched strong classical relaxations in the purely structural setting. We study whether learned representations can improve such solvers on one concrete problem: the combinatorial graph alignment problem (GAP), which seeks the node correspondence between two unlabeled graphs that maximizes the number of common edges. GAP encompasses graph isomorphism as a special case and, in its general form, reduces to the NP-hard quadratic assignment problem (QAP). The classical solver we build on is the Fast Approximate QAP algorithm ($\FAQ$) of \citet{vogelstein2015fast}, an existing Frank--Wolfe heuristic for a continuous relaxation of QAP whose output depends on its initialization (Section~\ref{sec:cont}). $\FAQ$ is not a contribution of this paper: our contributions are a corrected evaluation of it and a learned way to initialize it.
The graph matching problem can be cast as a GAP, and machine learning methods have been widely applied to it in pattern recognition \citep{conte2004thirty}, computer vision \citep{sun2020survey}, and social network analysis \citep{narayanan2008robust} (Section~\ref{sec:related}), motivated by the fact that in noisy real-world data the ground-truth matching may deviate from the mathematically optimal solution. We focus strictly on the combinatorial optimization setting, where only the optimal solution is relevant, and accordingly use the term \emph{graph alignment} rather than \emph{graph matching}.

\begin{table}[!t]
  \caption{Normalized alignment quality $\frac{\text{ALG}}{\text{REF}}$ (in \%) for sparse, dense and regular random graphs. $\Proj$ and $\FAQ$ are used to produce a permutation from the convex relaxation solution $D_\cx$ or from the similarity matrix computed by our chained FGNN (ChFGNN).}
\begin{center}
\begin{small}
\begin{sc}
  \begin{tabular}{llllll}
    \toprule
    type of graphs &  & sparse & dense & regular \\
    \midrule
    Baselines &$\Proj(D_\cx)$ & 17.3 & 24.4 & 2.9  \\
    (non-neural)& $\FAQ(D_\cx)$ & 67.1 & 53 & 27  \\
    \midrule
    Baselines
    & FGNN $\Proj$ & 17.8 & 23.6 & 6.7  \\
    (neural) & FGNN $\FAQ$ & 71.1 & 47 & 54  \\
    \midrule
    chaining & ChFGNN $\Proj$ & 95.8 & 44 & 67.1  \\
    chaining & ChFGNN $\FAQ$ & \bf{98.8} & \bf{77.4} & \bf{81.8}   \\
    \bottomrule
  \end{tabular}
\end{sc}
\end{small}
\label{tab:ratios}
\end{center}
\vskip -0.2in
\end{table}
\textbf{Iterative refinement through chaining.} We introduce \emph{chaining}: a sequential training procedure in which each GNN consumes a discrete ranking induced by the alignment decoded from the previous GNN. In our experiments, this ranking feedback produces similarity matrices that are substantially better initializations for $\FAQ$ on hard sparse and regular instances.
Table \ref{tab:ratios} illustrates our key results across different graph types.
To compare how close the different methods get to a common high-quality reference, we measure their \emph{normalized alignment quality}\footnote{In Table~\ref{tab:ratios}, $\mathrm{REF}$ is the same denominator for all methods in a column: sparse/dense Erd\H{o}s--R\'enyi at $p_{\noise}=0.25$ and regular graphs at $p_{\noise}=0.1$. It is a near-optimal reference score, not a certified optimum for every instance.} as
$\frac{\text{ALG}}{\text{REF}}$ in percent, where $\text{ALG}$ is the number of aligned edges obtained by the method and $\text{REF}$ is the corresponding reference score. Our chained GNNs, particularly when coupled with $\FAQ$ post-processing (ChFGNN $\FAQ$), consistently achieve the best performance.\\
We train and evaluate primarily on synthetic correlated random graphs --- the standard benchmarking methodology for combinatorial optimization \citep{skorin1990tabu, taillard1991robust} --- because they let us control problem difficulty via a single noise parameter and run fine-grained comparisons; we also evaluate on three real-world graph pairs (biology, social, road networks) to test transfer.\\
We formulate the graph alignment problem as a supervised learning task in Section \ref{sec:description}. While prior learning-based approaches have not matched $\FAQ(D_\cx)$ in the purely structural setting (Section \ref{sec:faq_sup}), we introduce a method that improves upon $\FAQ(D_\cx)$ on hard sparse and regular instances.
\textbf{Contributions.} (i) A corrected evaluation of $\FAQ$: initialized from the convex relaxation $D_\cx$ as advocated by \citet{lyzinski2015graph}, instead of the uniform matrix used in recent comparisons, it is a much stronger baseline than reported (Section~\ref{sec:faqinit}). (ii) Chaining, a training and inference procedure for learning $\FAQ$ initializations, with an FGNN backbone expressive enough to avoid the immediate 1-WL obstruction faced by constant-feature MPNNs on regular graphs. (iii) Comparisons on identical test pairs with the Gromov--Wasserstein solvers BAPG and FGWAlign, in addition to $\FAQ$ and FUGAL (Appendix~\ref{sec:previouser}). The empirical gains are largest on hard sparse Erd\H{o}s--R\'enyi graphs and on correlated regular graphs; dense ER and real-world results are more modest. 
We see iterative refinement with discrete combinatorial feedback as a promising direction for combinatorial problems with a natural ranking interpretation; we do not claim it transfers without modification to arbitrary CO problems.


\textbf{Mathematical notations.} Let $G=(V,E)$ be a simple graph with $V=\{1,\dots,n\}$ and adjacency matrix $A \in \{0,1\}^{n \times n}$, where $A_{ij}=1$ if $(i,j)\in E$ and $0$ otherwise. Let $\mathcal{S}_n$ denote the set of permutations of $V$, with each $\pi \in \mathcal{S}_n$ associated to a permutation matrix $P \in \{0,1\}^{n \times n}$ defined by $P_{ij}=1$ iff $\pi(i)=j$. The set of doubly stochastic matrices is denoted $\mathcal{D}_n$. For $A,B \in \mathbb{R}^{n \times n}$, the Frobenius inner product and norm are $\langle A,B\rangle = \mathrm{trace}(A^\top B)$ and $\|A\|_F = \sqrt{\langle A,A\rangle}$, respectively.

\section{From combinatorial optimization to a supervised learning task}\label{sec:description}

This section introduces the graph alignment problem (GAP) from a combinatorial optimization perspective, presents the $\FAQ$ algorithm, and describes how we formulate GAP as a learning problem using synthetic datasets with controllable difficulty.

\subsection{Graph alignment in combinatorial optimization}

\textbf{Problem formulation.} Given two $n\times n$ adjacency matrices $A$ and $B$ representing graphs $G_A$ and $G_B$, the graph alignment problem seeks to find the permutation that best aligns their structures. Formally, we minimize the Frobenius norm:
\begin{equation}
\label{eq:gm}\text{GAP}(A,B)  =\min_{P\in \mathcal{S}_n}\|AP-PB\|_F^2.
\end{equation}
Expanding the right-hand term, we see that minimizing \eqref{eq:gm} is equivalent to maximizing the number of matched edges:
\begin{equation}
\label{eq:qap}  \max_{P\in \mathcal{S}_n} \langle AP,PB\rangle= \max_{\pi \in \mathcal{S}_n}\sum_{i,j}A_{ij} B_{\pi(i)\pi(j)}.
\end{equation}
This formulation connects GAP to the broader class of Quadratic Assignment Problems (QAP) \citep{Burkard1998}.\\
\textbf{Computational complexity.} The GAP is NP-hard: when $G_B$ is a path or a cycle it is the Hamiltonian path or cycle problem, when $G_B$ is two disjoint cliques of size $n/2$ it is minimum bisection, and in general \eqref{eq:qap} is a maximum common subgraph problem, which is APX-hard \citep{crescenzi1995compendium}.\\
\textbf{Performance metrics.} We denote an optimal solution as $\pi^{A\to B}$. We evaluate alignment quality using two complementary metrics (that should be maximized), Accuracy and Number of common edges:
\begin{equation}
\label{eq:defmetrics}
\acc(\pi,\pi^{A\to B}) = \frac{1}{n} \sum_{i=1}^n \mathbf{1}(\pi(i) = \pi^{A\to B}(i)),
\qquad
\nce(\pi) = \frac{1}{2}\sum_{i,j}A_{ij} B_{\pi(i)\pi(j)}.
\end{equation}
Accuracy measures the fraction of correctly matched nodes, while the number of common edges quantifies structural similarity. In Table \ref{tab:ratios}, the ratio $\frac{\text{ALG}}{\text{REF}}$ is computed as $\frac{\nce(\pi^{\text{ALG}})}{\text{REF}}$, with the same reference score used for all methods in a graph family.
Note that even if this ratio is one, the accuracy may still be low if the GAP has no unique solution (as illustrated on real datasets in Section \ref{sec:realworld}).

\subsection{Continuous relaxations and the $\FAQ$ algorithm}\label{sec:cont}

Since the discrete optimization in \eqref{eq:gm} is intractable, continuous relaxations have been proposed where the discrete permutation set $\mathcal{S}_n$ is replaced by the continuous set of doubly stochastic matrices $\mathcal{D}_n$ in \eqref{eq:gm} or \eqref{eq:qap}:
\begin{itemize}
\item \textbf{Convex relaxation}:
\begin{align}
\label{eq:convrelax}
\arg\min_{D\in \mathcal{D}_n}\|AD-DB\|^2_F = D_{\text{cx}}
\end{align}
This yields a convex optimization problem with guaranteed global optimum.
\item \textbf{Indefinite relaxation}:
\begin{align}
\label{eq:indef}\max_{D\in \mathcal{D}_n}\langle AD, DB \rangle
\end{align}
This non-convex formulation often provides better solutions but is NP-hard in general due to its indefinite Hessian \citep{pardalos1991quadratic}.
\end{itemize}

\textbf{Solution extraction.} Both relaxations produce doubly stochastic matrices $D$ that must be projected to permutation matrices. This projection solves the linear assignment problem $\max_{P\in \mathcal{S}_n}\langle P, D\rangle$, efficiently solved by the Hungarian algorithm in $O(n^3)$ time \citep{khun1955}. We denote this projection as $\Proj(D) \in \mathcal{S}_n$.\\
\textbf{FAQ algorithm.} The Fast Approximate Quadratic ($\FAQ$) algorithm proposed by \cite{vogelstein2015fast} approximately solves the indefinite relaxation \eqref{eq:indef} using Frank-Wolfe optimization and then projects this solution in $\mathcal{S}_n$. Unlike the convex relaxation, $\FAQ$'s performance depends critically on initialization. We denote the $\FAQ$ solution with initial condition $D$ as $\FAQ(D) \in \mathcal{S}_n$.
As demonstrated in \cite{lyzinski2015graph}, $\FAQ$ often significantly outperforms simple projection: $\FAQ(D_{\text{cx}})$ typically yields much better solutions than $\Proj(D_{\text{cx}})$, especially for challenging instances. \textbf{This improvement motivates our approach of providing $\FAQ$ with better initializations through learned similarity matrices.}

\subsection{Synthetic datasets: controlled difficulty through noise}\label{sec:datasets}

\textbf{Connection to graph isomorphism.} When $G_A$ and $G_B$ are isomorphic (GAP$(A,B) = 0$), the alignment problem reduces to graph isomorphism, whose complexity remains open (neither known to be in P nor NP-complete; quasipolynomial by \cite{babai2016graph}). We study a natural generalization, noisy graph isomorphism, where a noise level controls the difficulty: at zero noise the graphs are isomorphic, and they become increasingly different as the noise grows.\\
\textbf{Correlated random graph model.} Our datasets consist of correlated random graph pairs $(G_A, G_B)$ with identical marginal distributions and controllable correlation: we generate correlated graphs $G_A$ and $G_B$ with known alignment, apply a random permutation $\pi^{\star} \in \mathcal{S}_n$ to $G_B$ to obtain $G_B'$, and use the triplets $(G_A, G_B', \pi^{\star})$ for supervised learning.\\
We employ three graph families—\textbf{Bernoulli}, \textbf{Erdős-Rényi}, and \textbf{Regular}—with parameters: \textbf{Number of nodes}: $n$; \textbf{Average degree}: $d$; \textbf{Noise level}: $p_{\noise} \in [0,1]$, see Section \ref{sec:def_graphs} for precise definitions.
The noise parameter controls edge correlation: the graphs $G_A$ and $G_B$ (before applying the random permutation) share $(1-p_{\noise})nd/2$ edges on average (with $p_{\noise} = 0$ yielding isomorphic graphs).
For low noise levels, we expect $\pi^{\star} = \pi^{A\to B}$, providing clean supervision. At high noise the planted permutation $\pi^{\star}$ may no longer be optimal, introducing label noise. We therefore report the number of common edges $\nce$ as our primary metric in addition to accuracy: $\nce$ is computed directly from the predicted alignment and the two graphs, and is \emph{independent} of $\pi^{\star}$. A method that finds a better alignment than the planted one simply attains a higher $\nce$.

\section{Learning via Chaining GNNs}\label{sec:chaining}

\textbf{Overview.} The chaining procedure iteratively refines alignment estimates through three operations: (1) computing node similarities, (2) extracting and scoring the current best permutation, and (3) using these scores to generate improved node features.

\subsection{Chaining procedure}

\textbf{Step 1: Initial feature extraction and similarity computation.}
Given a mapping $f$ that extracts node features from a graph's adjacency matrix $A\in \{0,1\}^{n\times n}$ and outputs $f:\{0,1\}^{n\times n}\rightarrow \R^{n\times d}$, we compute node feature matrices $f(A)$ and $f(B)$ for graphs $G_A$ and $G_B$. The initial similarity matrix captures pairwise node similarities via their feature dot products:
\begin{align}
    \label{eq:S0}S^{A\to B,(0)} = f(A)f(B)^T\in \R^{n\times n}.
\end{align}
Here, $S_{ij}^{A\to B, (0)}$ measures the similarity between node $i \in G_A$ and node $j \in G_B$ based on their learned features.

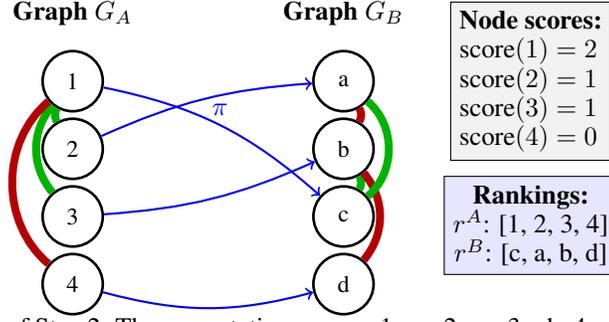
\begin{figure}[t]
  \centering
  \begin{tikzpicture}[
      scale=0.6, transform shape,
      node/.style={circle, draw=black, line width=1pt, minimum size=0.7cm, font=\scriptsize, fill=white, inner sep=1pt},
      arrow/.style={->, thick, blue},
      matched/.style={line width=2.5pt, green!70!black},
      unmatched/.style={line width=2.5pt, red!70!black}
  ]

  \node[node] (A1) at (0,2.5) {1};
  \node[node] (A2) at (0,1.5) {2};
  \node[node] (A3) at (0,0.5) {3};
  \node[node] (A4) at (0,-0.5) {4};

  \node[node] (B1) at (4,2.5) {a};
  \node[node] (B2) at (4,1.5) {b};
  \node[node] (B3) at (4,0.5) {c};
  \node[node] (B4) at (4,-0.5) {d};

 \draw[arrow] (A1) to[bend left=15] node[above, pos=0.5] {$\pi$} (B3);
 \draw[arrow] (A2) to[bend left=10] (B1);
 \draw[arrow] (A3) to[bend right=10] (B2);
 \draw[arrow] (A4) to[bend right=15] (B4);

  \draw[matched] (A1) to[bend right=30] (A2);
  \draw[matched] (A1) to[bend right=40] (A3);
  \draw[unmatched] (A1) to[bend right=50] (A4);

  \draw[unmatched] (B1) to[bend left=30] (B2);
  \draw[matched] (B2) to[bend left=30] (B3);
  \draw[unmatched] (B2) to[bend left=40] (B4);

  \draw[matched] (B1) to[bend left=50] (B3);

  \node[above=0.2cm of A1, font=\bfseries] {Graph $G_A$};
  \node[above=0.2cm of B1, font=\bfseries] {Graph $G_B$};

\node[draw, rectangle, right=1cm of B1, align=left, fill=gray!10] (scorebox) {
  \textbf{Node scores:} \\
  $\score(1) = 2$ \\
  $\score(2) = 1$ \\
  $\score(3) = 1$ \\
  $\score(4) = 0$
};

\node[draw, rectangle, below=0.2cm of scorebox, align=center, fill=blue!10] (ranking) {
  \textbf{Rankings:} \\
  $r^A$: [1, 2, 3, 4] \\
  $r^B$: [c, a, b, d]
};

  \end{tikzpicture}
  \caption{Illustration of Step 2. The permutation $\pi$ maps 1→c, 2→a, 3→b, 4→d. Green edges show matches: edge 1-2 with a-c, and edge 1-3 with b-c. Node 1 has the highest score (2 matched edges), nodes 2 and 3 each have 1 matched edge, and node 4 has no matched edges.}
  \label{fig:step2_illustration}
  \vskip -0.2in
\end{figure}

\textbf{Step 2: Permutation extraction and node quality scoring.}
From a similarity matrix $S^{A\to B}$, we extract the best permutation estimate by solving the linear assignment problem: $\pi=\Proj(S^{A\to B})$ where $\pi = \arg\max_{\pi\in \Scal_n} \sum_{i}S_{i \pi(i)}^{A\to B}$. This permutation $\pi:G_A\to G_B$ represents our current best guess for the optimal alignment $\pi^{A\to B}$.
To evaluate alignment quality, we compute a score for each node $i$ in graph $A$: $\score(i) = \sum_j A_{ij}B_{\pi(i)\pi(j)}$.
Intuitively, $\score(i)$ counts the number of edges incident to node $i$ that are correctly matched under the current permutation $\pi$---higher scores indicate better-aligned nodes (see Figure \ref{fig:step2_illustration}).
We then rank nodes in $G_A$ by decreasing score, obtaining a ranking $r^A\in \Scal_n$ such that:
\begin{align}
    \label{eq:ord}\score(r^A(1)) \geq \score(r^A(2)) \geq \dots \geq \score(r^A(n)).
\end{align}
We transfer this ordering to $G_B$ through the estimated permutation by setting $r^B(k)=\pi(r^A(k))$ for each rank $k$. Thus nodes with the same rank in $r^A$ and $r^B$ are matched by the current permutation: equivalently, $\pi(i)=r^B((r^A)^{-1}(i))$ (see Figure \ref{fig:step2_illustration}). The top-ranked pairs correspond to nodes whose incident edges are most consistent with the estimated alignment.

\textbf{Symmetry, ties, and the choice of score.} The score belongs to a matched \emph{pair}: computed from the side of $G_B$, $\sum_{y} B_{\pi(i)y}A_{i\pi^{-1}(y)} = \score(i)$, so both endpoints of a matched pair receive the same rank, and exchanging $G_A$ and $G_B$ (with $\pi\leftrightarrow\pi^{-1}$) yields the same rank fields up to the order of ties. Ties are frequent since scores are small integers; they are broken by the order returned by the sorting routine (\texttt{numpy.argsort}), which depends only on the score vector and uses no label information. We rank by the raw count rather than by a degree-normalized ratio because $\sum_i \score(i)=2\,\nce(\pi)$ is the per-node decomposition of the objective, and because the raw count promotes high-degree correctly matched nodes, which constrain the most neighboring assignments in the next stage; Section~\ref{sec:synthetic} shows that a chain retrained with the degree-normalized score performs on par, so this choice is second-order.

\textbf{Step 3: Ranking-enhanced feature learning.}
We now incorporate the ranking information to compute improved node features. Using a mapping $g:\{0,1\}^{n\times n}\times \Scal_n \rightarrow \R^{n\times d}$ that takes both the graph structure and node rankings as input, we compute enhanced feature matrices $g(A,r^A)$ and $g(B,r^B)$. The new similarity matrix is:
\begin{align}
    \label{eq:S1}S^{A\to B} = g(A,r^A)g(B,r^B)^T\in \R^{n\times n}.
\end{align}
This ranking-enhanced similarity matrix $S^{A\to B, (1)} = g(A,r^{A,(0)})g(B,r^{B,(0)})^T$ should be more informative than the initial $S^{A\to B,(0)}$ since it incorporates knowledge about which nodes align well. Consequently, we expect $\Proj(S^{A\to B,(1)})$ to be closer to the optimal $\pi^{A\to B}$ than $\Proj(S^{A\to B,(0)})$.

\textbf{What the ranking provides.} The next network does not reweight the previous similarities: it recomputes all $n^2$ pair similarities from $(A,r^A)$ and $(B,r^B)$ and solves a new assignment from scratch, so no match is frozen. What the ranking adds is evidence that did not exist before the projection. Matched pairs share a rank value across the two graphs, so a bilinear FGNN layer (Section~\ref{sec:architecture-main}) can count, for a candidate pair $(u,v)$, the neighbors of $u$ and of $v$ that carry the same rank value. For the true partner this is the number of common neighbors among correctly matched pairs; for a wrong candidate it is close to zero. The high-ranked matches thus serve as witnesses for re-assigning the low-ranked nodes, as in seeded graph alignment, where a partially correct correspondence is extended through one- and two-hop witnesses \citep{kazemi2015growing,lubars2018correcting,mossel2020seeded,yu2021graph}. Section~\ref{sec:synthetic} measures this correction directly.

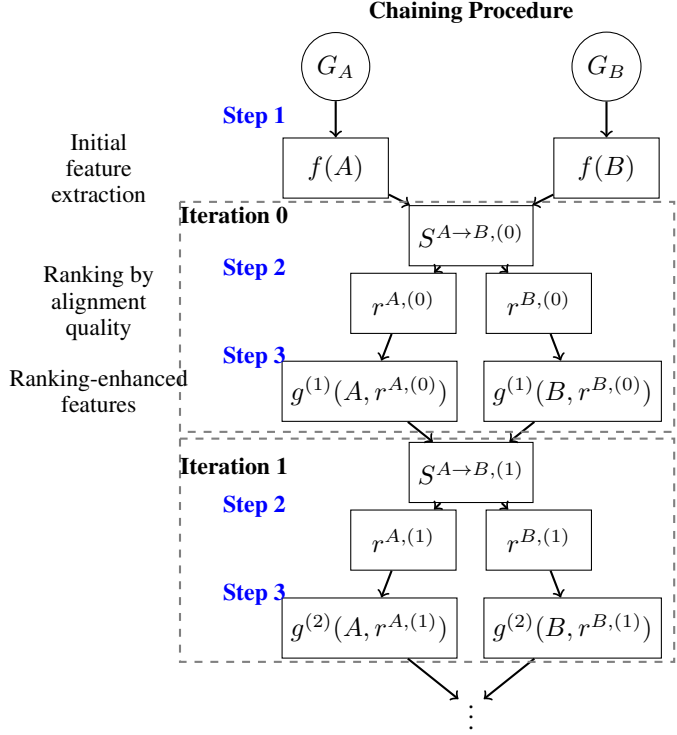
\begin{figure}[t]
  \centering
  \begin{tikzpicture}[
    scale=0.78, transform shape,
      box/.style={rectangle, draw, minimum width=1.1cm, minimum height=0.6cm, align=center, font=\footnotesize, inner sep=2pt},
      graph/.style={circle, draw, minimum size=0.6cm, align=center, font=\footnotesize, inner sep=1pt},
      arrow/.style={->, thick}
  ]

  \node[graph] (GA) at (0,0.8) {$G_A$};
  \node[graph] (GB) at (0,-0.8) {$G_B$};
  \node[box] (fA) at (1.6,0.8) {$f(A)$};
  \node[box] (fB) at (1.6,-0.8) {$f(B)$};
  \node[box] (S0) at (3.4,0) {$S^{A\to B,(0)}$};

  \node[box] (r0A) at (5.3,0.8) {$r^{A,(0)}$};
  \node[box] (r0B) at (5.3,-0.8) {$r^{B,(0)}$};

  \node[box] (g1A) at (7.5,0.8) {$g^{(1)}(A,r^{A,(0)})$};
  \node[box] (g1B) at (7.5,-0.8) {$g^{(1)}(B,r^{B,(0)})$};
  \node[box] (S1) at (10,0) {$S^{A\to B,(1)}$};

  \node (dots) at (11.6,0) {$\cdots$};
  \node[box] (SL) at (13.2,0) {$S^{A\to B,(L)}$};

  \draw[arrow] (GA) -- (fA);
  \draw[arrow] (GB) -- (fB);
  \draw[arrow] (fA) -- (S0);
  \draw[arrow] (fB) -- (S0);
  \draw[arrow] (S0) -- (r0A);
  \draw[arrow] (S0) -- (r0B);
  \draw[arrow] (r0A) -- (g1A);
  \draw[arrow] (r0B) -- (g1B);
  \draw[arrow] (g1A) -- (S1);
  \draw[arrow] (g1B) -- (S1);
  \draw[arrow] (S1) -- (dots);
  \draw[arrow] (dots) -- (SL);

  \node[blue, font=\scriptsize\bfseries] at (1.6,1.7) {Step 1};
  \node[blue, font=\scriptsize\bfseries] at (5.3,1.7) {Step 2};
  \node[blue, font=\scriptsize\bfseries] at (7.5,1.7) {Step 3};
  \node[blue, font=\scriptsize\bfseries, align=center] at (11.6,1.7) {repeat\\Steps 2,3};

  \end{tikzpicture}
  \caption{Overview of the chaining procedure. Starting from input graphs $G_A$ and $G_B$, we first (1) extract features and compute similarities, then iteratively (2) rank nodes by alignment quality, and (3) use rankings to enhance features and similarities.}
  \label{fig:chaining_overview}
  \vskip -0.2in
\end{figure}

\textbf{Iterative refinement.} We iterate steps 2 and 3 with different learned mappings $g^{(1)}, g^{(2)}, \ldots$ at each step, so that the chain of mappings $f,\, r,\, g^{(1)},\, g^{(2)},\, \ldots$ produces a sequence of similarity matrices $S^{A\to B,(0)},\, S^{A\to B,(1)},\, \ldots$ — each refining the previous via the discrete ranking signal. This creates a bootstrap effect where every step leverages the improved alignment from the previous one (Figure~\ref{fig:chaining_overview}).

\subsection{Training and inference with chained GNNs}\label{sec:training_chaining}

The ranking step $r$ is not differentiable, preventing end-to-end training. Instead, we train each GNN in the chain sequentially, which proves both practical and effective. This approach allows our method to explicitly learn from discrete permutation decisions at each step, which is crucial for the iterative improvement process.

\textbf{Sequential training procedure.} We implement $f$, $g^{(1)}, \ldots, g^{(k)}$ as graph neural networks and train them sequentially, each optimized to improve on the output of the previous stage. For training pairs $(G_A, G_B)$ with planted permutation $\pi^{\star}$, we use the cross-entropy loss
\begin{align}\label{eq:loss}
\mathcal{L}(S^{A\to B},\pi^{\star}) \;=\; -\sum_{i} \log \left(\mathrm{softmax}(S^{A\to B}_{i,\cdot})\right)_{\pi^{\star}(i)},
\end{align}
which encourages the similarity matrix to assign high mass to the correct correspondences. The full schedule is: train $f$ by minimizing $\mathcal{L}(S^{A\to B,(0)},\pi^{\star})$; then fix $f$, compute $r^{A,(0)}$ and $r^{B,(0)}$ on the training data and train $g^{(1)}$ by minimizing $\mathcal{L}(S^{A\to B,(1)},\pi^{\star})$; then fix $g^{(1)}$, compute $r^{A,(1)},r^{B,(1)}$ and train $g^{(2)}$ by minimizing $\mathcal{L}(S^{A\to B,(2)},\pi^{\star})$; and so on for $g^{(3)}, \ldots, g^{(k)}$.

This sequential approach ensures that each GNN learns to improve upon the alignment quality achieved by all previous networks in the chain.

\textbf{Why early errors do not compound.} First, every stage is trained on the \emph{actual} outputs of the previous stage on the training set, with the planted permutation as target: it sees at training time the same imperfect rankings that it receives at inference, including misassigned nodes, and the loss asks precisely for those to be corrected; this removes the train/test mismatch between stages that drives error accumulation in cascaded predictors \citep{ross2011reduction}. Second, nothing is frozen: the ranking is an input feature, not a constraint, and each stage re-solves the full assignment. Third, errors arrive labeled as unreliable, since a misassigned node preserves few incident edges and sits at the bottom of the ordering. Appendix~\ref{sec:ablations} measures the resulting correction at inference.

\textbf{Inference procedure.} On a new pair $(G_A, G_B)$, $f$ produces $S^{A\to B,(0)}$ and alternating applications of $r$ and $g^{(\ell)}$ yield refined similarity matrices $S^{A\to B,(1)}, \ldots, S^{A\to B,(L)}$. From any $S^{A\to B,(\ell)}$ we extract a candidate permutation $\pi^{(\ell)}$ via $\Proj$ or $\FAQ$ and rank candidates by $\nce(\pi^{(\ell)})$ from \eqref{eq:defmetrics}. Along the chain, accuracy and $\nce$ of $\pi^{(\ell)}$ increase monotonically, and the stages correct rather than preserve: of the nodes that $\pi^{(\ell)}$ misassigns and ranks in the bottom half, $\pi^{(\ell+1)}$ reassigns up to $35\%$ correctly, while it keeps $99\%$ or more of the correctly assigned top-half nodes from the third stage on (sparse graphs at noise $0.2$, Appendix~\ref{sec:ablations}).

\textbf{Looping at inference time.} At inference, we can apply the final trained network $g^{(L)}$ more than once: each ranking-then-refinement pass produces a new candidate alignment, and we keep the one with the highest $\nce$ seen so far. We call this \textbf{looping}. It introduces no additional training, and because we only keep an iterate when its $\nce$ improves on the running best, the returned alignment is guaranteed (in $\nce$) to be at least as good as the unlooped baseline. The rule only guarantees that looping cannot hurt. The gains come from re-applying a stage that was trained to correct its input. Section~\ref{sec:looping} shows that this gives substantial empirical gains on hard instances.

\subsection{GNN architecture and expressiveness}\label{sec:architecture-main}

\textbf{Architecture choice and motivation.} We implement all GNN mappings $f$, $g^{(1)}$, $g^{(2)}, \ldots, g^{(k)}$ using the same architecture, inspired by Folklore-type GNNs \citep{maron2019provably}. Unlike standard message-passing neural networks (MPNNs), this architecture operates on \emph{node pairs} rather than individual nodes, which is what gives it 2-FWL-level expressivity and lets it distinguish nodes in regular graphs that MPNNs cannot.

\textbf{Core architecture: Folklore-inspired residual layers.} Our GNN's main building block is a residual layer that processes hidden states for all node pairs $(h^{t}_{i\to j})_{i,j}\in \mathbb{R}^{n\times n\times d}$, producing updated states $(h^{t+1}_{i\to j})_{i,j}\in \mathbb{R}^{n\times n\times d}$:
$h^{t+1}_{i\to j} = h^{t}_{i\to j} + m_1\left( h^{t}_{i\to j}, \sum_\ell h^{t}_{i\to \ell} \odot m_0(h^{t}_{\ell\to j})\right)$,
where $m_0:\mathbb{R}^d \to \mathbb{R}^d$ and $m_1:\mathbb{R}^{2d} \to \mathbb{R}^d$ are multilayer perceptrons (MLPs) with graph normalization layers, and $\odot$ denotes component-wise multiplication. We refer to Section \ref{sec:architecture} for more details about our FGNN; in Section \ref{sec:expressivity} we prove that, despite the single-$m_0$ simplification, FGNN matches the 2-FWL expressivity of PPGN \citep{maron2019provably} up to a doubling of depth.

\section{Related work: state-of-the-art and learning limitations}\label{sec:faq_sup}

Additional related work on machine learning approaches to graph matching is discussed in Section \ref{sec:related}. In this section, we restrict our attention to the combinatorial optimization perspective.

\textbf{Non-learning methods.} Among traditional optimization approaches, $\FAQ$ represents the state-of-the-art for correlated random graphs \citep{lyzinski2015graph}, outperforming the convex relaxation, GLAG algorithm \citep{fiori2013robust}, PATH algorithm \citep{zaslavskiy2008path}, Umeyama's spectral method \citep{umeyama1988eigendecomposition}, and linear programming approaches \citep{almohamad1993linear}.
More recent papers \citep{xu2019scalable, bommakanti2024fugal} propose new algorithms for GAP, but their comparisons with $\FAQ$ underestimate it, likely because they use a suboptimal initialization (see Section~\ref{sec:faqinit}). Optimal-transport relaxations form another classical family: the Gromov--Wasserstein solvers SGWL \citep{xu2019scalable}, BAPG \citep{li2023convergent} and FGWAlign \citep{tang2025fused}. Evaluated on the same test pairs as all other methods (Section~\ref{sec:previouser}), BAPG and FGWAlign match $\FAQ(D_\cx)$ on dense graphs, fail earlier on sparse graphs, and are at chance accuracy on regular graphs.

\textbf{Learning approaches and their limitations.} Recent GNN-based methods for graph alignment include SeedGNN \citep{yu2023seedgnn}, PGM \citep{kazemi2015growing}, MGCN \citep{chen2020multi}, and MGNN \citep{wang2021neural}. The closest to our chaining procedure is Deep Graph Matching Consensus \citep{fey2020deep}, which also iteratively refines a similarity matrix, but with MPNN backbones and a differentiable consensus rule in the semantic setting where rich node features are available; our setting is purely structural and uses a non-differentiable ranking signal. For Erdős--Rényi graphs in our (purely structural) setting, none of the prior GNN methods reach positive accuracy at noise levels where $\FAQ(D_{\cx})$ still recovers a meaningful alignment (see Section~\ref{sec:previouser}).

These comparisons motivate $\FAQ(D_{\cx})$ as our main classical reference point in the purely structural correlated-graph setting. Our goal is not to replace $\FAQ$ but to provide it, through chaining, with similarity matrices that improve upon both the convex relaxation and prior learning approaches.

\begin{table}[!t]
  \caption{Accuracy ($\acc$) and number of common edges ($\nce$) for Erd\H{o}s-R\'enyi and regular graphs as a function of the noise $p_\noise$. Each cell is $\acc\,/\,\nce$. ChFGNN denotes our chained FGNNs; $\Proj$ and $\FAQ$ are post-processing decoders. The $\nce$ metric does not depend on the planted permutation $\pi^\star$ and is therefore unaffected by label imperfection at high noise.}
  \label{tab:ER-Reg}

\begin{center}
\resizebox{\columnwidth}{!}{%
\begin{sc}
\begin{tabular}{lcccccccc}
\toprule
\multicolumn{9}{c}{Sparse Erd\H{o}s-R\'enyi, average degree $4$ ($\nce_{\max}\!\approx\!1000$)}\\
\midrule
$p_\noise$ & 0 & 0.05 & 0.1 & 0.15 & 0.2 & 0.25 & 0.3 & 0.35 \\
\midrule
$\Proj(D_\cx)$  & \bf{0.98}/\bf{997} & \bf{0.97}/\bf{950} & 0.90/853 & 0.59/499 & 0.23/195 & 0.09/130 & 0.04/115 & 0.02/112 \\
$\FAQ(D_\cx)$   & \bf{0.98}/\bf{997} & \bf{0.98}/\bf{950} & \bf{0.96}/\bf{898} & \bf{0.95}/\bf{847} & 0.73/723 & 0.13/504 & 0.04/487 & 0.02/485 \\
\midrule
FGNN $\Proj$    & \bf{0.98}/\bf{997} & 0.94/\bf{925} & 0.74/674 & 0.44/365 & 0.23/193 & 0.12/134 & 0.06/114 & 0.03/103 \\
FGNN $\FAQ$     & \bf{0.98}/\bf{997} & \bf{0.98}/\bf{950} & \bf{0.96}/\bf{898} & \bf{0.95}/\bf{847} & 0.81/755 & 0.24/535 & 0.07/494 & 0.03/485 \\
\midrule
ChFGNN $\Proj$  & \bf{0.98}/\bf{997} & \bf{0.98}/\bf{950} & \bf{0.96}/\bf{898} & \bf{0.94}/\bf{845} & \bf{0.91}/\bf{790} & 0.82/\bf{720} & 0.49/549 & 0.08/367 \\
ChFGNN $\FAQ$   & \bf{0.98}/\bf{997} & \bf{0.98}/\bf{950} & \bf{0.96}/\bf{899} & \bf{0.95}/\bf{849} & \bf{0.93}/\bf{800} & \bf{0.85}/\bf{742} & 0.52/638 & 0.09/546 \\
\midrule
\multicolumn{9}{c}{Dense Erd\H{o}s-R\'enyi, average degree $80$ ($\nce_{\max}\!\approx\!20{,}000$)}\\
\midrule
$p_\noise$ & 0 & 0.05 & 0.1 & 0.15 & 0.2 & 0.25 & 0.3 & 0.35 \\
\midrule
$\Proj(D_\cx)$  & \bf{1.00}/\bf{19964} & \bf{1.00}/\bf{18987} & \bf{1.00}/\bf{17966} & 0.61/8700 & 0.14/3888 & 0.04/3646 & 0.02/3633 & 0.01/3624 \\
$\FAQ(D_\cx)$   & \bf{1.00}/\bf{19964} & \bf{1.00}/\bf{18987} & \bf{1.00}/\bf{17968} & \bf{1.00}/\bf{16990} & \bf{1.00}/\bf{15972} & 0.21/7922 & 0.01/6272 & 0.01/6276 \\
\midrule
FGNN $\Proj$    & \bf{1.00}/\bf{19964} & \bf{1.00}/\bf{18979} & 0.73/11254 & 0.28/4674 & 0.10/3651 & 0.04/3521 & 0.02/3517 & 0.01/3505 \\
FGNN $\FAQ$     & \bf{1.00}/\bf{19964} & \bf{1.00}/\bf{18987} & \bf{1.00}/\bf{17968} & \bf{1.00}/\bf{16990} & 0.95/15390 & 0.14/7031 & 0.01/6259 & 0.01/6254 \\
\midrule
ChFGNN $\Proj$  & \bf{1.00}/\bf{19964} & \bf{1.00}/\bf{18987} & 0.94/16241 & 0.83/13028 & 0.68/10291 & 0.37/6574 & 0.02/3690 & 0.01/3591 \\
ChFGNN $\FAQ$   & \bf{1.00}/\bf{19964} & \bf{1.00}/\bf{18987} & \bf{1.00}/\bf{17968} & \bf{1.00}/\bf{16990} & \bf{0.99}/\bf{15972} & \bf{0.62}/\bf{11577} & 0.01/6263 & 0.01/6255 \\
\midrule
\multicolumn{9}{c}{Regular random graphs, degree $10$ ($\nce_{\max}\!=\!2500$)}\\
\midrule
$p_\noise$ & \multicolumn{2}{c}{0} & \multicolumn{2}{c}{0.05} & \multicolumn{2}{c}{0.1} & \multicolumn{1}{c}{0.15} & 0.2\\
\midrule
$\Proj(D_\cx)$  & \multicolumn{2}{c}{0.002/51} & \multicolumn{2}{c}{0.002/51} & \multicolumn{2}{c}{0.003/50} & 0.001/49 & 0.002/50 \\
$\FAQ(D_\cx)$   & \multicolumn{2}{c}{0.002/385} & \multicolumn{2}{c}{0.003/425} & \multicolumn{2}{c}{0.003/456} & 0.002/369 & 0.003/496 \\
\midrule
FGNN $\Proj$    & \multicolumn{2}{c}{\bf{1.00}/\bf{2500}} & \multicolumn{2}{c}{0.31/405} & \multicolumn{2}{c}{0.03/113} & 0.005/108 & 0.003/106 \\
FGNN $\FAQ$     & \multicolumn{2}{c}{\bf{1.00}/\bf{2500}} & \multicolumn{2}{c}{\bf{0.95}/\bf{2059}} & \multicolumn{2}{c}{0.10/912} & 0.005/837 & 0.002/838 \\
\midrule
ChFGNN $\Proj$  & \multicolumn{2}{c}{\bf{1.00}/\bf{2500}} & \multicolumn{2}{c}{\bf{0.95}/\bf{2034}} & \multicolumn{2}{c}{0.54/1135} & 0.009/281 & 0.003/95 \\
ChFGNN $\FAQ$   & \multicolumn{2}{c}{\bf{1.00}/\bf{2500}} & \multicolumn{2}{c}{\bf{0.95}/\bf{2059}} & \multicolumn{2}{c}{\bf{0.56}/\bf{1383}} & 0.008/871 & 0.003/836 \\
\bottomrule
\end{tabular}
\end{sc}}
\end{center}
\vskip -0.2in
\end{table}

\section{Empirical results and comparison to $\FAQ$}\label{sec:results}

We evaluate our chaining procedure against $\FAQ$ \cite{vogelstein2015fast}, the strongest method on all our datasets (Sections \ref{sec:previousreal} and \ref{sec:previouser} compare with more recent algorithms, including the Gromov--Wasserstein solvers BAPG and FGWAlign). We compare non-neural baselines using the convex relaxation against our chained FGNNs (i.e.\ $f, g^{(1)},\dots , g^{(L)}$); an architecture ablation against single-step MPNN backbones (SAGE, GIN, GAT) appears in Table~\ref{tab:arch-ablation}. All methods can be combined with $\Proj$ and $\FAQ$ as a post-processing step to extract a permutation (Section \ref{sec:cont}).

\subsection{Main results on synthetic datasets}\label{sec:synthetic}

Table \ref{tab:ER-Reg} presents results on graphs with $500$ nodes across noise levels, for sparse Erd\H{o}s-R\'enyi graphs (average degree 4), dense Erd\H{o}s-R\'enyi graphs (average degree 80), and regular graphs (degree 10); higher $p_{\text{noise}}$ means more corrupted alignments.

\textbf{Sparse and dense Erdős-Rényi graphs.} For both sparse and dense graphs, our chained FGNNs significantly outperform all baselines, particularly at challenging noise levels. At $p_{\text{noise}} = 0.25$, chained FGNNs with $\FAQ$ post-processing achieve $85\%$ accuracy on sparse graphs, compared to $13\%$ for the non-neural $\FAQ$ baseline.
Note that $p_\noise{=}0.2$ corresponds to the setting of \cite{yu2023seedgnn} where none of the GNN-based methods they evaluate achieve positive accuracy (see Section~\ref{sec:previouser}); the architecture ablation in Table~\ref{tab:arch-ablation} confirms that single-step MPNN backbones (SAGE, GIN, GAT) likewise collapse in this regime, while chaining with our FGNN backbone reaches $0.93$.

\textbf{Regular graphs: a particularly challenging case.} Regular graphs are a regime where standard approaches break. The convex relaxation \eqref{eq:convrelax} admits the barycenter $D_{\text{cx}} = J := \frac{1}{n}\mathbf{1}\mathbf{1}^\top$ as a solution; the Frank--Wolfe gradient of $\langle AD, DB\rangle$ at $D = J$ equals $2d^2 J$ on a $d$-regular pair (since $AJ = JB = dJ$), so the linear-assignment subproblem at the first $\FAQ$ iteration is degenerate --- every permutation matrix achieves the same inner product --- and $\FAQ$'s trajectory from this initialisation is determined by the LAP solver's tie-breaking rather than by the graph data. MPNNs face a separate expressivity obstruction: bounded by 1-WL \citep{xu2018powerful}, they collapse to identical node embeddings on a regular graph with constant initial features, even though a random $d$-regular graph has trivial automorphism group almost surely, so their similarity matrices are uninformative not because of symmetry but because 1-WL fails to refine. Table~\ref{tab:ER-Reg} shows that only our chained FGNN gets a meaningful result ($56\%$ accuracy at $p_{\text{noise}} = 0.1$, where every other method essentially fails); both the 2-FWL backbone (expressivity beyond 1-WL) and the chaining (refinement of the noisy alignment) are needed, an empirical gap that we do not formally bound.

\textbf{Why the gains are larger on sparse graphs.} The ranking relies on two sources of contrast, and both fade as the graph densifies. The first is degree heterogeneity: in an Erd\H{o}s--R\'enyi graph of average degree $d$, degrees vary by a relative amount of about $1/\sqrt d$, so sparse graphs offer many nodes that stand out for the first stage, while dense graphs approach the regular case where none does. The second is the evidence that an already-matched neighbor pair gives about a candidate pair: it supports the true candidate with probability $\lambda(1-p_\noise)$ but a wrong one with probability $\lambda^2$, where $\lambda=d/n$, a ratio $(1-p_\noise)/\lambda$ that falls as $d$ grows. On dense graphs the transition to failure is therefore sharper and the noise window in which chaining improves on $\FAQ(D_\cx)$ is narrower (Appendix~\ref{sec:ablations}). 

\textbf{Ranking ablations.} Table~\ref{tab:rank-ablation} (Appendix~\ref{sec:ablations}) isolates what the ranking contributes, on the sparse chain trained at $p_\noise=0.22$ with $\FAQ$ decoding on $30$ fixed test pairs per noise level. \emph{Removing} the rank field reduces the chain to independently trained single FGNNs, so the gap between the ChFGNN and FGNN rows of Table~\ref{tab:ER-Reg} measures the contribution of the ranking. Ranking nodes by $\score(i)/\deg(i)$ instead of $\score(i)$ and retraining the chain gives comparable accuracies at every noise level ($0.924$ against $0.928$ at $p_\noise=0.2$, $0.840$ against $0.884$ at $0.25$), so the normalization is a second-order choice. Applying the normalized ranking at inference to the chain trained with the raw score, without retraining, collapses accuracy at $p_\noise\geq0.2$: each stage relies on the ordering convention it was trained with. A \emph{random order} keeps the transported matching (matched nodes still share a rank value) but destroys the confidence information: at inference it is catastrophic ($0.346$ at $p_\noise=0.2$), and a chain retrained end-to-end on random orders still improves stage over stage but never catches the score-ordered chain, with a gap that widens with noise ($0.802$ against $0.928$ at $0.2$, $0.331$ against $0.884$ at $0.25$). The confidence ordering is thus the component that keeps the chain accurate in the regime where chaining matters. Per-stage measurements (Appendix~\ref{sec:ablations}) show that the stages correct rather than propagate errors: at test noise $0.2$, the probability that a bottom-half misassigned node is corrected by the next stage rises from $0.14$ to $0.35$ as the reliable set grows, while top-half correct matches are retained with probability $0.99$--$1.00$ from the third stage on.

\textbf{Architecture ablation: chaining versus 2-FWL expressiveness.}
To disentangle the contribution of the FGNN backbone from the chaining procedure, we re-train the entire pipeline with three standard message-passing GNN backbones in place of FGNN: SAGE, GIN, and GAT. All backbones use the same hidden width and number of residual blocks. Table \ref{tab:arch-ablation} reports both metrics on sparse Erd\H{o}s-R\'enyi graphs ($n{=}500$, $d{=}4$) with $\FAQ$ decoding ($\Proj$ decoding gives the same ordering). Two findings stand out: \emph{(i)} chaining improves every backbone, e.g.\ ChGAT$\to$0.92 vs.\ GAT$\to$0.14 at $p_\noise{=}0.15$; \emph{(ii)} 2-FWL expressiveness is needed in the harder regime --- only ChFGNN maintains accuracy at $p_\noise\geq 0.20$. On dense Erd\H{o}s-R\'enyi and regular graphs we were unable to successfully train any of the MPNN backbones: regular graphs are theoretically beyond MPNN expressivity \cite{xu2018powerful}.

\begin{table}[!t]
  \caption{Architecture ablation on sparse Erd\H{o}s-R\'enyi ($n{=}500$, $d{=}4$), $\FAQ$ decoding. Each cell is $\acc\,/\,\nce$. Ch$\cdot$ denotes the chained variant of each backbone. Chaining improves every backbone; only ChFGNN survives at $p_\noise\geq 0.20$.}
  \label{tab:arch-ablation}
\begin{center}
\begin{small}
\begin{sc}
\begin{tabular}{lcccccc}
\toprule
$p_\noise$ & 0 & 0.05 & 0.1 & 0.15 & 0.2 & 0.25 \\
\midrule
SAGE-$\FAQ$    & \bf{0.98}/\bf{997} & \bf{0.97}/\bf{946} & 0.13/497 & 0.05/477 & 0.02/474 & 0.01/473 \\
ChSAGE-$\FAQ$  & \bf{0.98}/\bf{997} & \bf{0.97}/\bf{951} & 0.69/756 & 0.11/494 & 0.04/476 & 0.02/475 \\
\midrule
GIN-$\FAQ$     & \bf{0.98}/\bf{997} & \bf{0.98}/\bf{951} & 0.93/\bf{879} & 0.25/538 & 0.04/478 & 0.02/477 \\
ChGIN-$\FAQ$   & \bf{0.98}/\bf{997} & \bf{0.98}/\bf{951} & \bf{0.95}/\bf{889} & 0.34/575 & 0.06/483 & 0.03/477 \\
\midrule
GAT-$\FAQ$     & \bf{0.98}/\bf{997} & \bf{0.97}/\bf{951} & 0.79/805 & 0.14/503 & 0.04/479 & 0.02/478 \\
ChGAT-$\FAQ$   & \bf{0.98}/\bf{997} & \bf{0.97}/\bf{951} & \bf{0.96}/\bf{893} & 0.92/\bf{833} & 0.24/537 & 0.05/486 \\
\midrule
ChFGNN-$\FAQ$  & \bf{0.98}/\bf{997} & \bf{0.98}/\bf{950} & \bf{0.96}/\bf{899} & \bf{0.95}/\bf{849} & \bf{0.93}/\bf{800} & \bf{0.85}/\bf{730} \\
\bottomrule
\end{tabular}
\end{sc}
\end{small}
\end{center}
\vskip -0.3in
\end{table}

\textbf{Training strategy: noise-level selection.} Figure~\ref{fig:noise} (appendix) shows that intermediate training noise levels (around $p_{\text{noise}} = 0.22$ for sparse graphs) generalize best: too-easy instances do not generalize to harder cases and too-hard instances are suboptimal on easier ones. The training noise of each graph family is chosen on the validation split only, before evaluating on the test pairs of Table~\ref{tab:ER-Reg}; the appendix tables sweep training noise levels without this selection and, unless stated, without the final $\nce$-based inference loop.

\textbf{Looping: enhanced inference without additional training.}\label{sec:looping}
The last network $g^{(L)}$ can be re-applied at inference for free. Since it was supervised on the output of $g^{(L-1)}$, further applications are formally out of distribution, so we treat looping as an empirically grounded heuristic with a safety net: after each pass we recompute $\nce$ and keep the candidate only if it improves the running best (capped at $N_{\text{loop}}{=}100$), so that the returned alignment is at least as good (in $\nce$) as the unlooped one. Table~\ref{tab:accuracy_comparison} (appendix) shows substantial gains on hard instances (up to $+0.26$ accuracy at $p_\text{noise}{=}0.30$, $L+1{=}16$) at negligible compute overhead; the number of loops scales with difficulty (avg.\ 15 at $p_\text{noise}{=}0.15$ vs.\ 91 at $0.30$).

\subsection{Computational cost and scalability}\label{sec:compute}
A fair hardware-independent comparison of running times between $\FAQ(D_{\cx})$ and our chained GNN procedure is challenging, so we focus on inference complexity and iteration counts rather than claiming algorithmic speedups from wall-clock time. While our method requires an initial GPU-based training phase, this is assumed to be completed before solving new instances.
For $\FAQ(D_{\cx})$, each gradient step involves solving a linear assignment problem ($O(n^3)$), and total runtime depends on the number of gradient ascent iterations.
Our chaining procedure has two main costs as $n$ grows: ($i$) the bilinear pair aggregation in the graph layer, with arithmetic cost $O(n^3F)$ and activation storage $O(n^2F)$ per layer when implemented as matrix products/tensor contractions, where $F$ is the feature dimension of the pair state ($F=256$ in our models); and ($ii$) computing ranks via a projection $\Proj$ of the similarity matrix in each iteration, an $O(n^3)$ CPU operation.
Table~\ref{tab:iter} (appendix) (appendix) reports the average number of gradient ascent iterations in $\FAQ$, starting from either $D_\cx$ or the similarity matrix produced by our chained FGNN. The iteration count is substantially lower with the chained FGNN, indicating that the similarity matrix from chaining provides a more accurate initialization than $D_\cx$.
A wall-clock time--performance trade-off against $\FAQ(J)$, $\FAQ(D_\cx)$ and FUGAL (Pareto curves) is reported in Appendix~\ref{sec:timing}.
\textbf{Memory is the binding constraint.} The order-2 pair state ($0.26$\,GB per stored activation at $n=500$, $F=256$, single precision) grows quadratically with $n$. The released implementation runs end-to-end on the largest real graphs of this paper ($n=1004$ and $n=1174$), but this dense implementation, does not exploit graph sparsity, preventing our method to scale to graphs orders of magnitude larger. 

\begin{table}[!t]
  \caption{Accuracy ($\acc$) and number of common edges ($\nce$) on noisy versions of real-world networks; each cell is $\acc\,/\,\nce$. ChFGNN-ER4 is the model trained on sparse Erd\H{o}s--R\'enyi graphs ($d{=}4$); ChFGNN is trained on the target network and noise level.}
  \label{tab:realworld-noisy}
\begin{center}
\begin{small}
\begin{sc}
  \begin{tabular}{lcccccc}
    \toprule
   Method & \multicolumn{2}{c}{yeast25LC} & \multicolumn{2}{c}{ca-netscience} & \multicolumn{2}{c}{inf-euroroad} \\
   \cmidrule(lr){2-3} \cmidrule(lr){4-5} \cmidrule(lr){6-7}
   Noise $q$ & 5\% & 10\% & 10\% & 20\% & 10\% & 20\% \\
    \midrule
      $\FAQ(D_\cx)$ & 0.50/7660 & 0.45/7245 & 0.65/822 & 0.46/687 & 0.56/1170 & 0.11/940 \\
    \midrule
    ChFGNN-ER4 & 0.48/7693 & 0.42/7297 & 0.64/818 & 0.44/688 & 0.40/1111 & 0.08/\bf{970} \\
    \midrule
    ChFGNN & 0.54/7732 & 0.51/\bf{7404} & 0.65/824 & 0.57/\bf{724} & 0.64/\bf{1213} & 0.15/\bf{963} \\
    \bottomrule
  \end{tabular}
  \end{sc}
\end{small}
\end{center}
\vskip -0.2in
\end{table}

\subsection{Results on real graphs.}\label{sec:realworld} We evaluate on three real-world datasets (biology, social networks and road networks; details in Section \ref{sec:previousreal}). \textbf{Yeast} \citep{10.1109/TCBB.2017.2740381} is a protein--protein interaction (PPI) network (1{,}004 proteins, 8{,}323 trusted interactions) with five noisy variants obtained by adding $q\%$ low-confidence edges ($q \in \{5,10,15,20,25\}$). Since the base graph is always an induced subgraph of each variant, the maximum number of common edges is $8{,}323$ and the true node correspondence is known. In Table~\ref{tab:multimagna-full} (appendix), all methods recover nearly the maximum number of common edges (within 3\%), but high $\nce$ does not always translate into high node accuracy because the base graph has a large automorphism group; $\nce$ is thus the more reliable metric, and $\FAQ$ is already near-optimal on this benchmark.
For harder benchmarks, we apply the edge-addition--removal noise model (Section~\ref{sec:datasets}) to the yeast PPI with $q=25\%$, the \textbf{ca-netscience} coauthorship network \citep{PhysRevE.74.036104}, and the \textbf{inf-euroroad} road network \citep{vsubelj2011robust}, evaluating both zero-shot transferred and dataset-specific FGNNs. Table~\ref{tab:realworld-noisy} shows that dataset-specific ChFGNNs achieve the best $\nce$ in most high-noise settings, improving over $\FAQ(D_\cx)$ by about 2\%. The transferred ChFGNN-ER4 model is more mixed: it sometimes improves $\nce$ but underperforms $\FAQ(D_\cx)$ in node accuracy, so results for zero-shot transfer are limited. The smaller gains have two causes. Distribution shift: the real graphs are far from the synthetic training model in degree spread and clustering (Table~\ref{tab:realworld-structure}, appendix), so transfer is reliable only when the target resembles the training distribution, as for ca-netscience at $10\%$ noise where the transferred model is within one point of the dataset-specific one, and dataset-specific training, which only requires simulating the noise model on the target graph without external labels, is otherwise needed. Evaluation ambiguity: on ca-netscience at $10\%$ noise the chain reaches $\nce=824$ against $826$ for $\FAQ$ started from the true alignment (Table~\ref{tab:realworld-noisy-full}), yet its node accuracy is $0.65$. The accuracy is limited by symmetries of the graph: $48\%$ of the nodes of ca-netscience have an exact twin, a node with the same neighbors (typically a coauthor of the same papers), and exchanging twins changes no $\nce$, so the expected accuracy of any method that only sees the structure is at most $0.71$ on this graph. Exact twins alone bound the accuracy by $0.85$ on yeast and $0.98$ on inf-euroroad; there the limiting factor is the near-twins, nodes that can be exchanged with another node by changing at most two edges of the graph, which the noise makes indistinguishable in practice: $57\%$ of the nodes of yeast and $77\%$ of those of inf-euroroad are in this case (Table~\ref{tab:realworld-structure}). We therefore report both metrics and do not read much into accuracy differences of a few points.

\section{Conclusion and limitations}\label{sec:conclusion}

We introduced a chained-GNN procedure for combinatorial graph alignment that learns initializations for $\FAQ$ and improves over $\FAQ(D_\cx)$ on hard sparse correlated Erd\H{o}s--R\'enyi instances and on the regular-graph regime among the methods we evaluate, including Gromov--Wasserstein solvers; its ablations identify the confidence ordering of the ranking as the component that makes the chain reach useful accuracy in a feasible number of stages. It extends naturally to the seeded GAP. \textbf{Limitations:} each FGNN layer has $O(n^2F)$ activation memory, capping training around $n\!\approx\!1000$ on an 80\,GB GPU in our implementation, and the dense pair state cannot exploit graph sparsity; real-world gains over a properly initialized $\FAQ$ are consistent across all six settings but smaller than on synthetic graphs, and they require training on the target graph rather than zero-shot transfer; FGNN inherits the 2-FWL ceiling \citep{maron2019provably} and cannot separate strongly-regular graphs of identical parameters, a barrier we have not hit empirically.

\begin{ack}
We thank the anonymous reviewers and the area chair, whose questions on the mechanism of the ranking feedback, on robustness and on baselines led to the analyses of Section~\ref{sec:synthetic} and Appendix~\ref{sec:ablations}. We are grateful to the CLEPS infrastructure from Inria Paris for providing resources and support. This work was granted access to the HPC resources of IDRIS under the allocation 2025-AD010613995R2 and A0191016874 made by GENCI, on the A100 and H100 partitions of the Jean Zay supercomputer.
\end{ack}

\bibliography{references}
\bibliographystyle{plainnat}

\newpage
\appendix

\section{Appendix}

\subsection{Correlated random graphs}\label{sec:def_graphs}

In this section, we present the mathematical details for the various correlated random graphs model used in this paper.

\paragraph*{Bernoulli graphs.}

We start with the model considered in \citep{lyzinski2015graph}.
Given $n$ the number of nodes, $\rho\in [0,1]$ and a symmetric hollow matrix $\Lambda\in [0,1]^{n\times n}$, define $\Ecal=\{\{i,j\}, i\in [n], j\in [n], i\neq j\}$. Two random graphs $G_A=(V_A,E_A)$ and $G_B=(V_B,E_B)$ are $\rho$-correlated Bernoulli($\Lambda$) distributed, if for all $\{i,j\}\in \Ecal$, the random variables (matrix entries) $A_{ij}$ and $B_{ij}$ are such that $B_{ij}\sim \Bernoulli(\Lambda_{ij})$ independently drawn and then conditioning on $B$, we have $A_{ij}\sim \Bernoulli(\rho B_{ij} + (1-\rho)\Lambda_{ij})$ independently drawn. Note that the marginal distributions of $A$ and $B$ are both Bernoulli($\Lambda$), so the laws of $A$ and $B$ are the same (but correlated).

In our experiments in Sections \ref{sec:training} and \ref{sec:inference}, we consider the same case as in \citep{lyzinski2015graph}: $n=150$ vertices, the entries of the matrix $\Lambda$ are i.i.d.\ uniform in $[\alpha, 1-\alpha]$ with $\alpha=0.1$, and we vary $\rho$.

\paragraph*{Erd\H{o}s-R\'enyi graphs.}

The Erd\H{o}s-R\'enyi model is a special case of the Bernoulli model where $\Lambda$ is the matrix with all entries equal to $\lambda$. To be consistent with the main notation, we define $p_{\noise} = (1-\lambda)(1-\rho)$ where $\rho$ was the correlation above and $\lambda = d/n$ where $d$ is the average degree of the graph. Hence the random graphs $G_A$ and $G_B$ are correlated Erd\H{o}s-R\'enyi graphs when $\Pb(A_{i,j}=B_{i,j}=1)=\frac{d}{n}(1-p_{\noise})$ and $\Pb(A_{i,j}=0, \: B_{i,j}=1)= \Pb(A_{i,j}=1, \: B_{i,j}=0) = \frac{d}{n}p_{\noise}$.

\paragraph*{Regular graphs.}

In this case, we first generate $G_A$ as a uniform regular graph with degree $d$ and then we generate $G_B$ by applying edgeswap to $G_A$: if $\{i, j\}$ and $\{k, \ell\}$ are two edges of $G_A$ then we swap them to $\{i,\ell\}$ and $\{k, j\}$ with probability $p_{\noise}$.

\begin{table}[b!]
  \centering
  \caption{Statistics of synthetic datasets.}
\begin{tabular}{lllll}
  \toprule
    name & average & number  & used for comparison & sizes of \\
     & degree & of nodes & with& train/val\\
  \midrule
  Bernoulli &  70 & 150 & $\FAQ(D_\cx)$ \citep{lyzinski2015graph}&2000/200\\
  \midrule
  Sparse Erd\H{o}s-R\'enyi (ER 4) &4 & 500 & MPNN \citep{yu2023seedgnn} &200/100\\
  \midrule
  Dense Erd\H{o}s-R\'enyi (ER 80) & 80& 500 & MPNN \citep{yu2023seedgnn}& 200/100 \\
  \midrule
  Large Erd\H{o}s-R\'enyi & 3 & 1000 &  Bayesian message passing  & 200/100\\
  &&& \citep{muratori2024faster} & \\
  \midrule
  Regular & 10 & 500 & new& 200/100\\
\bottomrule
      \end{tabular}

\end{table}

\subsection{Limitations in Reported $\FAQ$ Performance in Prior Work}\label{sec:faqinit}

Several recent studies on graph alignment do not provide a fair or representative comparison against the Fast Approximate Quadratic Assignment ($\FAQ$) algorithm \citep{vogelstein2015fast}. In particular, $\FAQ$'s performance is frequently underestimated due to the use of suboptimal initialization strategies. When properly initialized, $\FAQ$ remains a remarkably strong and competitive baseline.

The original $\FAQ$ implementation is provided in MATLAB in the authors' repository\footnote{\url{https://github.com/jovo/FastApproximateQAP/}}
 and dates back nearly a decade. A more recent and widely used implementation is available in SciPy via
\texttt{scipy.optimize.quadratic\_assignment(method='faq')}. While convenient, the SciPy implementation does not expose the theoretically motivated initialization scheme recommended in the literature.

Specifically, \citep{lyzinski2015graph} shows that initializing the indefinite relaxation (i.e., $\FAQ$) with the solution of the convex relaxation $D_\cx$, leads to improved convergence and empirical performance. Their theoretical analysis and experiments both support this strategy.

The corresponding MATLAB implementation computes this initialization using a Frank–Wolfe procedure (see \texttt{relaxed\_normAPPB\_FW\_seeds.m} in the original repository). In contrast, SciPy's default behavior initializes FAQ with the uniform doubly stochastic matrix, which is known to be inferior to the convex-optimum initialization.

To ensure a fair evaluation, we therefore explicitly compute the convex relaxation using our own Frank–Wolfe solver and pass the resulting matrix $D_\cx$ as the initialization for $\FAQ$. This matches the procedure advocated in \citep{lyzinski2015graph} and the original MATLAB implementation.

Below, we empirically assessed the impact of initialization on three real-world datasets (biology, social networks, and transportation) previously used in \citep{bommakanti2024fugal} to introduce the algorithm FUGAL. Across all datasets, we observe that:
\begin{enumerate}
  \item Our $\FAQ$ results differ substantially from those reported in \citep{bommakanti2024fugal};
  \item Properly initialized $\FAQ$ consistently outperforms FUGAL and all competing methods reported in \cite{bommakanti2024fugal}; and
  \item Our chained FGNN models further outperform this strengthened $\FAQ$ baseline.
\end{enumerate}

We also re-evaluated FAQ on the synthetic benchmarks used in SeedGNN \citep{yu2023seedgnn}. Although SeedGNN compares against PGM \citep{kazemi2015growing}, SGM \citep{fishkind2019seeded}, and MGCN \citep{chen2020multi}, it does not include $\FAQ$. Using the convex-optimum initialization, we find that $\FAQ$ outperforms SeedGNN and all of its reported competitors, while our chained FGNNs again achieve the best overall performance.

Taken together, these results indicate that prior studies likely underestimate $\FAQ$ due to inadequate initialization, and that properly configured $\FAQ$ should be regarded as a strong and necessary baseline for graph alignment research.

\begin{table}[t!]
\caption{Summary statistics of the real-world graphs.}
\label{tab:realworld_stats}
\centering
\begin{tabular}{lcccc}
\toprule
\textbf{Dataset} & \textbf{\# Nodes} & \textbf{\# Edges} & \textbf{Avg. Degree} & \textbf{size train/valid}\\
\midrule
Yeast PPI \citep{10.1109/TCBB.2017.2740381} & 1,004 & 8,323 & 16.58 & 20/20\\
ca-netscience\citep{PhysRevE.74.036104} & 379 & 914 & 4.82 & 200/20\\
inf-euroroad \citep{vsubelj2011robust} & 1,174 & 1,417 & 2.41 & 20/5\\
\bottomrule
\end{tabular}
\end{table}

\begin{table}[t!]
\caption{Structure and symmetries of the real-world graphs. Degree spread is the standard deviation of the degrees divided by their mean, and clustering the average clustering coefficient; both are compared with an Erd\H{o}s--R\'enyi graph of the same size and average degree. A node has a \emph{twin} when another node has the same neighbors (or, for two adjacent nodes, the same other neighbors): exchanging twins is an automorphism of the graph, so it leaves $\nce$ unchanged for every alignment, and the accuracy ceiling is the expected accuracy of a method that cannot tell twins apart (one correct node per twin class). The last column counts the nodes that can be exchanged with another node at the cost of at most two edges of the graph.}
\label{tab:realworld-structure}
\centering
\begin{small}
\setlength{\tabcolsep}{5pt}
\begin{tabular}{lccccc}
\toprule
\textbf{Dataset} & \textbf{Degree spread} & \textbf{Clustering} & \textbf{Twins} & \textbf{Acc.\ ceiling} & \textbf{Swap $\leq 2$ edges}\\
 & graph / ER & graph / ER & & & \\
\midrule
Yeast PPI & 1.12 / 0.24 & 0.65 / 0.02 & 24\% & 0.85 & 57\%\\
ca-netscience & 0.81 / 0.45 & 0.74 / 0.01 & 48\% & 0.71 & 83\%\\
inf-euroroad & 0.49 / 0.64 & 0.02 / 0.002 & 4\% & 0.98 & 77\%\\
\bottomrule
\end{tabular}
\end{small}
\end{table}

\begin{table}[b!]

  \caption{Accuracy ($\acc$) and number of common edges ($\nce$) on the Yeast PPI networks. ChFGNN ER4 is our model trained on Erd\H{o}s-R\'enyi graphs with average degree 4, while ChFGNN is trained on the pairs obtained with the three first networks.}
  \label{tab:multimagna-full}
\begin{center}
\begin{small}
\begin{sc}
  Yeast PPI networks (acc / nce ) \\
  \begin{tabular}{lllllll}
    \toprule
   {Method} & 5\% conf & 10\% conf & 15\% conf & 20\% conf& 25\% conf \\
    \midrule
    $\FAQ(J)$ & 37.5 / 7383 & 34.4 / 7245 & 29.1 / 6807 & 23.9 / 6689 & 36.4 / 7383 \\
    SGWL  & 83.6 / -- & -- / -- & 66.6 / -- & -- / -- & 58.8 / -- \\
    FUGAL  & 83.0 / 8311 & 77.7 / 8231 & 74.3 / 8172 & 70.9 / 8148 & 68.6 / 8095 \\
     $\FAQ(D_\cx)$ & 84.2 / 8323 & 82.6 / 8317 & 78.0 / 8289 & 77.0 / 8294 & 76.1 / 8306 \\
    \midrule
    ChFGNN ER4 & 80.3 / 8300 & 75.3 / 8288 & 67.2 / 8252 & 63.1 / 8213 & 53.1 / 8080 \\
    \midrule
    ChFGNN & training & training & training & 72.2 / 8300 & 69.8 / 8291 \\
    \bottomrule
  \end{tabular}
  \end{sc}
\end{small}
\end{center}
\end{table}

\subsection{Recent results on real graphs}\label{sec:previousreal}

The real-world networks in Section~\ref{sec:realworld} are standard benchmarks for graph alignment. We apply the same noising procedure as for Erdős--Rényi graphs (described above), using the original graph's average degree~\(d\). Table \ref{tab:realworld_stats} gives the sizes of the training and validation sets used for training our chained FGNNs.

Table~\ref{tab:multimagna-full} reports results for SGWL~\citep{xu2019scalable} and FUGAL~\citep{bommakanti2024fugal}. SGWL results come from the original paper; FUGAL results were obtained using the authors' code (available at \url{https://github.com/idea-iitd/Fugal}). Our FUGAL performance matches~\citep{bommakanti2024fugal}, but our \(\FAQ\) results do not. We attribute the discrepancy to poor initialization: using the uninformative barycenter \(J = \frac{1}{n}\mathbf{1}\mathbf{1}^{\top}\) reproduces the degraded \(\FAQ\) performance reported in~\citep{bommakanti2024fugal}.

On noisy real datasets (Table~\ref{tab:realworld-noisy-full}), FUGAL never matches \(\FAQ\), contrary to the claims in~\citep{bommakanti2024fugal}. To compute the maximum number of common edges, we run \(\FAQ\) initialized with the true permutation (prior to noising).

\begin{table}[!t]

  \caption{Accuracy ($\acc$) and number of common edges ($\nce$) on noisy versions of real-world networks. Each network is corrupted by adding noise at different levels. ChFGNN ER4 is trained on Erd\H{o}s-R\'enyi graphs, while ChFGNN is trained on the specific network and noise level. In bold if gain in $\nce$ is larger than 2\%.}
  \label{tab:realworld-noisy-full}
\begin{center}
\begin{small}
\begin{sc}
  Real-world networks with added noise (acc / nce) \\
  \begin{tabular}{llcccccc}
    \toprule
   {Method} & \multicolumn{2}{c}{yeast25LC} & \multicolumn{2}{c}{ca-netscience} & \multicolumn{2}{c}{inf-euroroad} \\
   \cmidrule(lr){2-3} \cmidrule(lr){4-5} \cmidrule(lr){6-7}
   & 5\% & 10\% & 10\% & 20\% & 10\% & 20\% \\
    \midrule
    FUGAL & 53.1 / 7480 & 44.6 / 7035 & 60.3 / 794 & 37.7 / 629 & 18.3 / 818 & 2.9 / 714 \\
      $\FAQ(D_\cx)$ & 49.8 / 7660 & 44.7 / 7245 & 65.2 / 822 & 45.6 / 687 & 55.8 / 1170 & 10.9 / 940 \\
    \midrule
    ChFGNN ER4 & 47.6 / 7693 & 42.3 / 7297 & 63.5 / 818 & 44.1 / 688 & 40.0 / 1111 & 7.5 / \bf{970} \\
    \midrule
    ChFGNN & 54.1 / 7732 & 51.3 / \bf{7404} & 65.4 / 824 & 57.0 / \bf{724} & 63.5 / \bf{1213} & 15.4 / \bf{963} \\
    \midrule
    Max $\nce$ & 7909 & 7498 & 826 & 730 & 1272 & 1137 \\
    \bottomrule
  \end{tabular}
  \end{sc}
\end{small}
\end{center}
\end{table}

\subsection{More recent results on synthetic graphs}\label{sec:previouser}

Because FUGAL \citep{bommakanti2024fugal} reports stronger performance than SGWL \citep{xu2019scalable}, we evaluated FUGAL on our synthetic benchmarks using the authors' public implementation (\url{https://github.com/idea-iitd/Fugal}). We tested multiple hyperparameter configurations to ensure a fair comparison. Nevertheless, across all settings we considered, FUGAL consistently underperformed both $\FAQ$ (with convex-optimum initialization $D_{\mathrm{cx}}$) and our chained FGNN (ChFGNN) models.

Table~\ref{tab:fugal_synth_nce} reports the number of correctly matched common edges ($\nce$). Properly initialized $\FAQ$ substantially outperforms FUGAL in all regimes, while ChFGNN achieves the best overall performance.

\begin{table}[h]
\centering
\caption{Number of common edges (NCE) recovered on synthetic graphs.}
\label{tab:fugal_synth_nce}
\begin{tabular}{lccc}
\toprule
\textbf{Datasets} & \textbf{FUGAL} & $\FAQ(D_{\mathrm{cx}})$ & \textbf{ChFGNN} \\
\midrule
ER sparse (noise $0.1$) & 560 & 898 & 899 \\
ER sparse (noise $0.2$) & 396 & 723 & 800 \\
ER dense (noise $0.1$) & 17968 & 17968 & 17968 \\
ER dense (noise $0.2$) & 5853 & 15972 & 15876 \\
Regular (noise $0.05$) & 131 & 425 & 2059 \\
\bottomrule
\end{tabular}
\end{table}

\paragraph{Gromov--Wasserstein solvers: BAPG and FGWAlign.}
We also evaluate two optimal-transport baselines: BAPG \citep{li2023convergent}, a single-loop Bregman alternating projected gradient method for the Gromov--Wasserstein relaxation, and FGWAlign \citep{tang2025fused}, a fused Gromov--Wasserstein solver designed for graph edit distance. For equal-size unlabeled graphs, the graph edit distance minimized by FGWAlign equals $|E_A|+|E_B|-2\,\nce(\pi)$ over permutations, so it directly optimizes our metric. BAPG follows the protocol of its authors through its implementation in the POT library \citep{flamary2021pot}: raw adjacency matrices, uniform marginals, step size $\rho=0.1$, $2000$ iterations, tolerance $10^{-6}$, square loss; we decode its transport plan with the Hungarian algorithm, a stronger extraction than the row-wise argmax used in the original paper. FGWAlign uses the authors' code with their default full protocol (patience $15$, top-$k$ $5$) in its sparse mode, since the dense code path overflows single precision at $n=500$; the two modes have the same optimum over permutations, since every edge or non-edge mismatch count is an affine function of $\nce(\pi)$ once the two edge counts are fixed. Both solvers are run on exactly the same test pairs as every other method ($30$ pairs per cell for sparse and regular graphs, $10$ for dense), so that the comparison is paired per graph pair; the per-pair results are released with the code.

\begin{figure}[t]
  \centering
  \includegraphics[width=\textwidth]{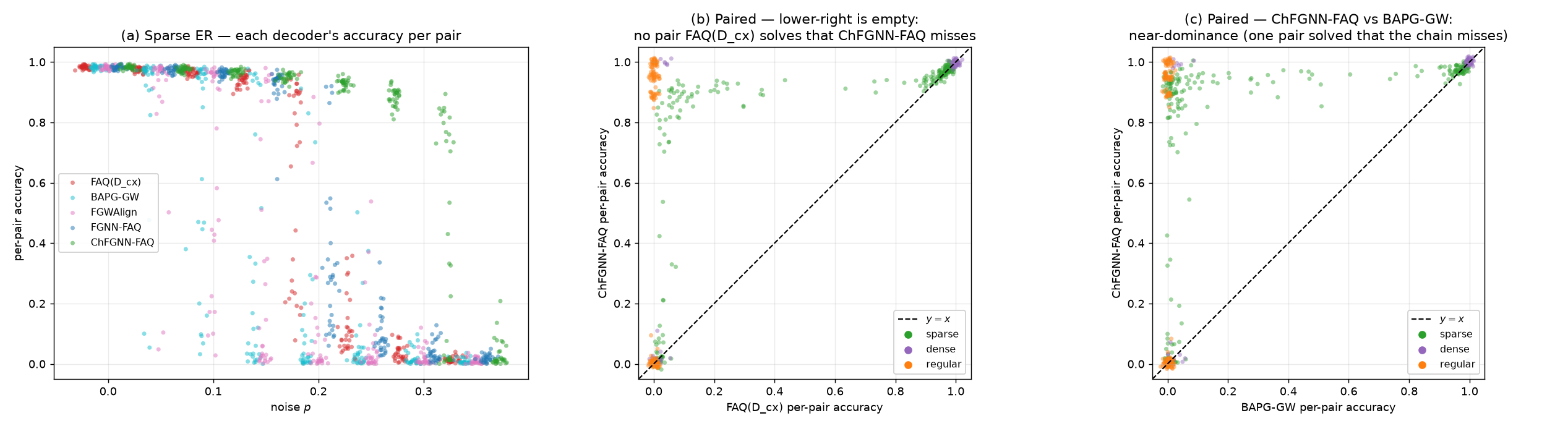}
  \caption{Per-pair analysis on the test pairs. (a) Per-pair accuracy of each decoder on sparse Erd\H{o}s--R\'enyi graphs as a function of the noise: the transition is bimodal (pairs are either solved or not) and the two Gromov--Wasserstein solvers transition earliest. (b) ChFGNN $\FAQ$ against $\FAQ(D_\cx)$ on every synthetic pair: the empty lower-right region means that no pair solved by the baseline is missed by the chain. (c) The same paired view against BAPG: a single pair (dense, $p_\noise=0.3$) is solved by BAPG and missed by the chain.}
  \label{fig:per_sample}
\end{figure}

Table~\ref{tab:otgw} reports accuracy and $\nce$ for all methods on these pairs, reproduced with the released code. The means differ slightly from Table~\ref{tab:ER-Reg}, which used $100$ pairs per cell from another dataset. BAPG and FGWAlign behave as one: calling a pair solved when accuracy exceeds $0.5$, the two solvers solve or fail the same pairs on $469$ of the $470$ synthetic pairs ($240$ sparse, $80$ dense, $150$ regular). On dense graphs they coincide with $\FAQ(D_\cx)$ and collapse at the same noise level; on sparse graphs they degrade earlier than $\FAQ(D_\cx)$; on regular graphs both are at chance accuracy, like every relaxation-based method we tested. FGWAlign does salvage more common edges on regular graphs ($\nce\approx 820$ of $2500$, against $\approx 50$ for BAPG and $\Proj(D_\cx)$) without recovering node identities. Across all $470$ pairs, ChFGNN $\FAQ$ solves $192$ pairs that BAPG does not and misses a single pair that BAPG solves (dense, $p_\noise=0.3$); Figure~\ref{fig:per_sample} shows the paired, per-pair view, together with the bimodality of all methods at their transition.

\begin{table}[!t]
  \caption{Accuracy ($\acc$) and number of common edges ($\nce$) of the Gromov--Wasserstein solvers BAPG and FGWAlign, compared with the convex-relaxation baselines and our chained FGNNs on identical seeded test pairs ($n=500$). Each cell is $\acc\,/\,\nce$; best accuracy per column in bold.}
  \label{tab:otgw}
\begin{center}
\resizebox{\columnwidth}{!}{%
\begin{sc}
\begin{tabular}{lcccccccc}
\toprule
\multicolumn{9}{c}{Sparse Erd\H{o}s-R\'enyi, average degree $4$ ($30$ pairs per cell)}\\
\midrule
$p_\noise$ & 0 & 0.05 & 0.1 & 0.15 & 0.2 & 0.25 & 0.3 & 0.35 \\
\midrule
$\Proj(D_\cx)$ & \bf{0.98}/994 & \bf{0.98}/944 & 0.91/849 & 0.58/482 & 0.22/195 & 0.088/131 & 0.040/117 & 0.020/117 \\
$\FAQ(D_\cx)$ & \bf{0.98}/994 & \bf{0.97}/945 & \bf{0.97}/895 & 0.94/841 & 0.64/683 & 0.12/499 & 0.037/484 & 0.015/481 \\
\midrule
BAPG & \bf{0.98}/994 & 0.89/893 & 0.71/772 & 0.35/624 & 0.13/548 & 0.062/528 & 0.020/520 & 0.009/520 \\
FGWAlign & \bf{0.98}/994 & 0.88/886 & 0.70/767 & 0.34/618 & 0.13/546 & 0.062/528 & 0.021/520 & 0.010/520 \\
\midrule
ChFGNN $\Proj$ & \bf{0.98}/994 & \bf{0.98}/945 & \bf{0.97}/894 & \bf{0.95}/840 & 0.91/783 & 0.85/722 & 0.40/451 & 0.035/261 \\
ChFGNN $\FAQ$ & \bf{0.98}/994 & \bf{0.98}/945 & \bf{0.97}/895 & \bf{0.95}/843 & \bf{0.93}/792 & \bf{0.88}/744 & \bf{0.44}/608 & 0.033/520 \\
\midrule
\multicolumn{9}{c}{Dense Erd\H{o}s-R\'enyi, average degree $80$ ($10$ pairs per cell)}\\
\midrule
$p_\noise$ & 0 & 0.05 & 0.1 & 0.15 & 0.2 & 0.25 & 0.3 & 0.35 \\
\midrule
$\Proj(D_\cx)$ & \bf{1.00}/19906 & \bf{1.00}/18904 & \bf{1.00}/17882 & 0.67/9827 & 0.17/3975 & 0.049/3664 & 0.024/3646 & 0.012/3604 \\
$\FAQ(D_\cx)$ & \bf{1.00}/19906 & \bf{1.00}/18904 & \bf{1.00}/17896 & \bf{1.00}/16912 & \bf{1.00}/15899 & 0.32/8854 & 0.014/6226 & 0.006/6218 \\
\midrule
BAPG & \bf{1.00}/19906 & \bf{1.00}/18904 & \bf{1.00}/17896 & \bf{1.00}/16912 & \bf{1.00}/15899 & 0.32/8832 & 0.13/7018 & 0.011/6208 \\
FGWAlign & \bf{1.00}/19906 & \bf{1.00}/18904 & \bf{1.00}/17869 & \bf{1.00}/16842 & \bf{1.00}/15784 & 0.32/8595 & 0.13/6981 & 0.014/6208 \\
\midrule
ChFGNN $\Proj$ & \bf{1.00}/19906 & \bf{1.00}/18904 & 0.95/16584 & 0.82/12699 & 0.73/10333 & 0.49/7026 & 0.082/4278 & 0.014/4037 \\
ChFGNN $\FAQ$ & \bf{1.00}/19906 & \bf{1.00}/18904 & \bf{1.00}/17896 & \bf{1.00}/16912 & \bf{1.00}/15899 & \bf{0.90}/14050 & \bf{0.22}/7800 & 0.009/6272 \\
\midrule
\multicolumn{9}{c}{Regular random graphs, degree $10$ ($30$ pairs per cell)}\\
\midrule
$p_\noise$ & \multicolumn{2}{c}{0} & \multicolumn{2}{c}{0.05} & \multicolumn{2}{c}{0.1} & 0.15 & 0.2 \\
\midrule
$\Proj(D_\cx)$ & \multicolumn{2}{c}{0.002/51} & \multicolumn{2}{c}{0.002/51} & \multicolumn{2}{c}{0.002/51} & 0.002/50 & 0.002/50 \\
$\FAQ(D_\cx)$ & \multicolumn{2}{c}{0.002/623} & \multicolumn{2}{c}{0.002/491} & \multicolumn{2}{c}{0.002/571} & 0.002/646 & 0.002/409 \\
\midrule
BAPG & \multicolumn{2}{c}{0.002/51} & \multicolumn{2}{c}{0.002/51} & \multicolumn{2}{c}{0.002/51} & 0.002/50 & 0.002/50 \\
FGWAlign & \multicolumn{2}{c}{0.003/821} & \multicolumn{2}{c}{0.002/820} & \multicolumn{2}{c}{0.003/821} & 0.002/821 & 0.002/821 \\
\midrule
ChFGNN $\Proj$ & \multicolumn{2}{c}{\bf{1.00}/2500} & \multicolumn{2}{c}{0.72/1329} & \multicolumn{2}{c}{0.27/518} & 0.005/296 & 0.004/290 \\
ChFGNN $\FAQ$ & \multicolumn{2}{c}{\bf{1.00}/2500} & \multicolumn{2}{c}{\bf{0.96}/2052} & \multicolumn{2}{c}{\bf{0.60}/1416} & 0.003/865 & 0.003/869 \\
\bottomrule
\end{tabular}
\end{sc}}
\end{center}
\end{table}

The problem of graph alignment for correlated Erdős R\'enyi random graphs has been studied empirically with Message Passing GNN (MPNN) in \citep{yu2023seedgnn} when a seed of matched vertices is given in addition to the 2 graphs. We are reproducing their results taken from \url{https://github.com/Leron33/SeedGNN} corresponding to Figure 6 in \citep{yu2023seedgnn}. SeedGNN refers to \citep{yu2023seedgnn}, PGM to \citep{kazemi2015growing}, SGM to \citep{fishkind2019seeded} and MGCN to \citep{chen2020multi}.

At $0\%$ seeds (the purely structural setting studied in this paper) and noise $0.2$, none of the four seeded methods in Table~\ref{tab:seed} exceeds $0.4\%$ accuracy on either sparse or dense graphs. By contrast, on the same regime, $\FAQ(D_\cx)$ from Section~\ref{sec:results} reaches $73\%$ (sparse) and $100\%$ (dense), and our chained FGNNs reach $93\%$ and $99\%$ respectively.

\begin{table}[t!]
\centering
\caption{Accuracy (\%) on sparse (average degree 4) and dense (average degree 80) Erdős R\'enyi random graphs at noise $0.2$, as a function of the seed fraction. Note the different seed scales for the two regimes.}
\label{tab:seed}
\begin{tabular}{lccccccccccc}
\toprule
\multicolumn{12}{c}{\textit{Sparse, average degree 4}} \\
\midrule
\textbf{Fraction of Seeds} & 0\% & 2\% & 4\% & 6\% & 8\% & 10\% & 12\% & 14\% & 16\% & 18\% & 20\% \\
\midrule
SeedGNN & 0.3 & 15.1 & 47.4 & 82.8 & 96.0 & 96.6 & 97.0 & 97.6 & 97.6 & 97.6 & 97.6 \\
PGM & 0.2 & 2.3 & 6.1 & 16.3 & 31.6 & 54.5 & 73.3 & 79.2 & 86.3 & 88.9 & 92.7 \\
SGM & 0.3 & 3.6 & 8.9 & 13.8 & 22.3 & 36.3 & 54.5 & 67.3 & 84.4 & 89.6 & 91.6 \\
MGCN & 0.1 & 2.0 & 4.0 & 6.7 & 8.4 & 11.1 & 12.4 & 14.0 & 16.3 & 18.9 & 20.5 \\
\midrule
\multicolumn{12}{c}{\textit{Dense, average degree 80}} \\
\midrule
\textbf{Fraction of Seeds} & 0\% & 0.5\% & 1\% & 1.5\% & 2\% & 2.5\% & 3\% & 3.5\% & 4\% & 4.5\% & 5\% \\
\midrule
SeedGNN & 0.1 & 0.7 & 91.4 & 100 & 100 & 100 & 100 & 100 & 100 & 100 & 100 \\
PGM & 0.1 & 0.6 & 1.8 & 4.3 & 19.3 & 51.2 & 96.6 & 100 & 100 & 100 & 100 \\
SGM & 0.2 & 1.5 & 85.8 & 100 & 100 & 100 & 100 & 100 & 100 & 100 & 100 \\
MGCN & 0.1 & 0.7 & 1.5 & 1.9 & 3.7 & 5.2 & 6.9 & 8.0 & 10.9 & 12.3 & 13.7 \\
\bottomrule
\end{tabular}
\end{table}

\subsection{Time--performance trade-off}\label{sec:timing}

\begin{figure}[h]
  \begin{center}
  \includegraphics[width=0.65\textwidth]{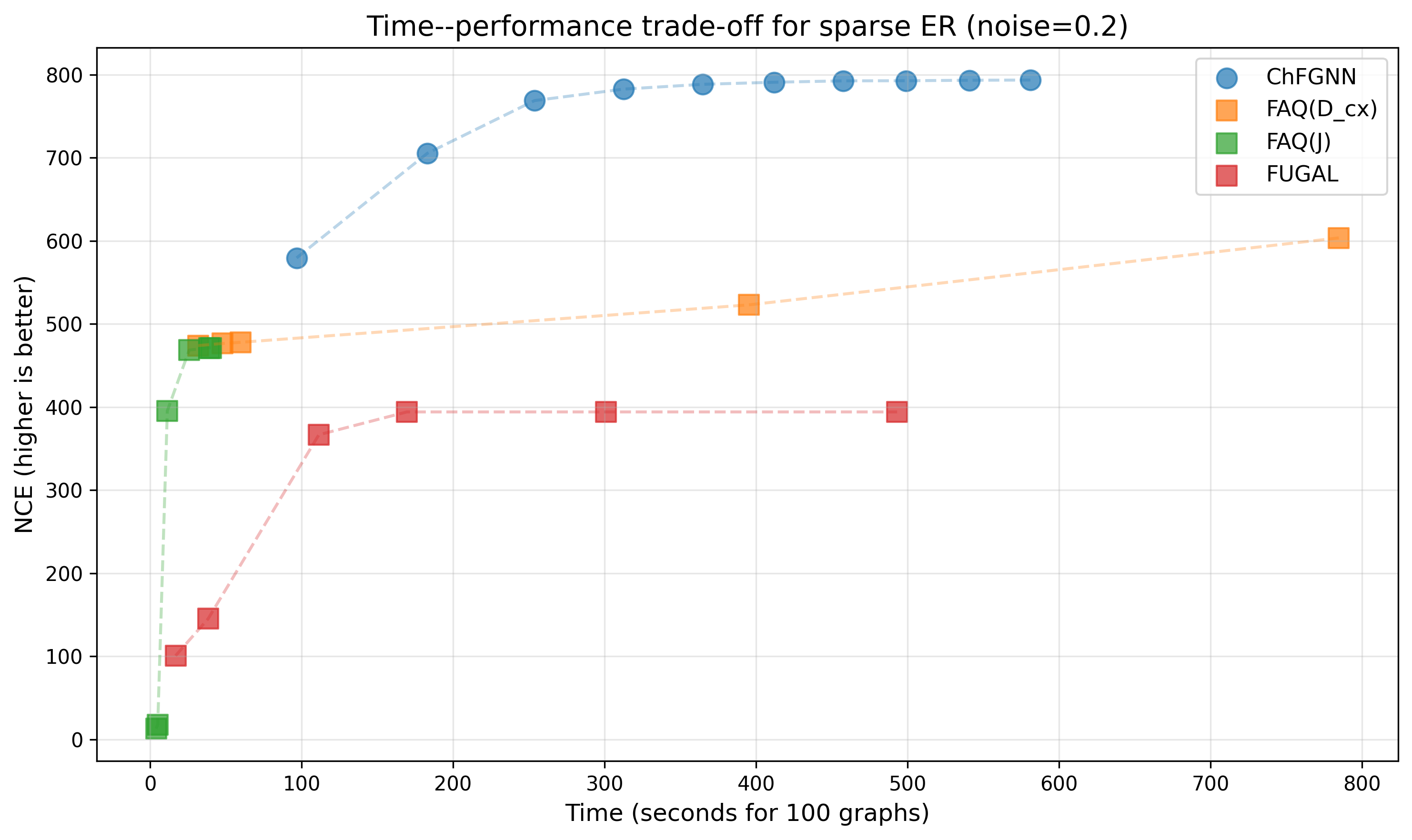}
  \caption{Number of common edges ($\nce$) recovered in sparse Erd\H{o}s--R\'enyi graphs as a function of wall-clock computation time allocated to each implementation: FUGAL and $\FAQ$ variants run on CPU, while ChFGNN runs on GPU. The plot should be read as an implementation-level trade-off, not as a hardware-independent speed comparison.
}
  \label{fig:pareto}
  \end{center}
\end{figure}

For iterative algorithms, the computation time can be controlled by adjusting the number of iterations. This applies to gradient-based methods such as the Frank--Wolfe algorithm, used to compute either the convex relaxation $D_{\mathrm{cx}}$ or the indefinite relaxation $\FAQ$. The recently proposed FUGAL algorithm \citep{bommakanti2024fugal} is also iterative. Finally, our chained FGNNs naturally define an iterative procedure, where we may bound the number of chaining steps.
Figure~\ref{fig:pareto} displays the resulting Pareto curves, showing the trade-off between performance and runtime for each method on sparse Erd\H{o}s--R\'enyi graphs with noise level $p_{\noise} = 0.2$.

For FUGAL, we used the authors' implementation (available at \url{https://github.com/idea-iitd/Fugal}), specifically the \texttt{predict\_alignment} routine with hyperparameter \texttt{mu = 1}.
For $\FAQ(J)$, we relied on the SciPy implementation \texttt{scipy.optimize.quadratic\_assignment(method='faq')}. For $\FAQ(D_{\mathrm{cx}})$, we used our own Frank--Wolfe implementation to compute $D_{\mathrm{cx}}$ before passing it to $\FAQ$.

We emphasize that $\FAQ$ and FUGAL run on CPU, whereas our chained FGNNs run on GPU. Runtimes correspond to computing the number of common edges ($\nce$) over 100 graph pairs of size $n = 500$. Thus Figure~\ref{fig:pareto} is an implementation-level wall-clock comparison, not a hardware-agnostic efficiency claim. The more robust conclusion is that the chained FGNN reaches higher $\nce$ with a small number of learned refinement steps, while $\FAQ(D_\cx)$ remains sensitive to the number and implementation quality of Frank--Wolfe iterations.

FUGAL is substantially faster than \(\FAQ(D_\cx)\). We lack an efficient implementation of the Frank--Wolfe solver required for the convex-relaxation initialization \(D_\cx\), and our implementation prioritizes correctness over speed. The subsequent \(\FAQ\) step uses SciPy's efficient routine \texttt{quadratic\_assignment}. We expect that \(\FAQ(D_\cx)\) could be made faster with an optimized implementation.

\paragraph{Per-pair runtimes of the baselines.} Table~\ref{tab:runtime} complements the Pareto curves with per-pair inference times measured on a single CPU core, for the Gromov--Wasserstein solvers of Section~\ref{sec:previouser} and, on the same laptop, for $\FAQ(D_\cx)$ (dominated by our unoptimized Frank--Wolfe solver for $D_\cx$) and for the full chain run on CPU. None of the baselines has a training phase; the chain's training is a one-off cost discussed in Section~\ref{sec:compute}. For BAPG the wall-clock time of every pair is stored with the released results and averaged over the test pairs of each family; for FGWAlign only the per-cell averages printed during the runs were kept, and the table reports their range across noise levels.
The memory footprint of all these methods is $O(n^2)$ (dense transport plans or cost products for the optimal-transport solvers, dense doubly stochastic iterates for $\FAQ$); the chain's constant is larger by the feature dimension of the pair state ($256$ in our models, Section~\ref{sec:compute}).

\begin{table}[h]
\centering
\caption{Inference time per graph pair at $n=500$ on a single laptop CPU core (seconds). The GPU timings of the chain are given by the Pareto curves of Figure~\ref{fig:pareto} (a few seconds per pair).}
\label{tab:runtime}
\small
\begin{tabular}{lccc}
\toprule
Method & sparse ($d{=}4$) & dense ($d{=}80$) & regular ($d{=}10$) \\
\midrule
BAPG & 7 & 3 & $<0.1$ \\
FGWAlign & $\approx 70$ & $150$--$210$ & $85$--$95$ \\
$\FAQ(D_\cx)$ & $20$--$40$ & -- & -- \\
ChFGNN $\FAQ$ (full chain, CPU only) & $\approx 90$ & -- & -- \\
\bottomrule
\end{tabular}
\end{table}

Table~\ref{tab:iter} reports the average number of Frank--Wolfe iterations of $\FAQ$ on sparse Erd\H{o}s--R\'enyi graphs when it is started from $D_\cx$ or from the similarity matrix of the chained FGNN: the learned initialization substantially lowers the number of iterations.

\begin{table}[h]
  \caption{Average number of gradient projections ($\Proj$) in the Frank-Wolfe algorithm $\FAQ$, with initialization from either $D_\cx$ or the similarity matrix produced by the chained FGNNs.}
  \label{tab:iter}
\begin{center}
\begin{small}
\begin{sc}
  \begin{tabular}{llllll}
    \toprule
   ER 4 - noise & 0.15  & 0.2 & 0.25 & 0.3 & 0.35 \\
    \midrule
    $\FAQ(D_\cx)$&  15.6 &	31.4	& 25.8	& 24.2	& 25.6 \\
    ChFGNN $\FAQ$ &  4.1	& 6.5	& 8.3	& 15.1	& 19.7 \\
    \bottomrule
  \end{tabular}
\end{sc}
\end{small}
\end{center}
\end{table}

\subsection{GNN architecture and expressiveness}\label{sec:architecture}

The choice of a more expressive architecture is crucial for our approach. The success of the chaining procedure critically depends on producing a high-quality initial similarity matrix $S^{A\to B}(0)$ to bootstrap the iterative refinement process. Standard MPNNs, which aggregate only local neighborhood information, would produce similarity matrices based on limited local features—insufficient for capturing the global structural patterns needed for effective graph alignment. This limitation has been observed in prior work: \cite{nowak2018revised} implemented a similar initial step using MPNNs with limited success, while \cite{azizian2021expressive} demonstrated the superiority of Folklore-type GNNs for this task. As we show in Section \ref{sec:results}, combining our expressive architecture with the chaining procedure yields substantial performance improvements over single-step approaches.

\textbf{Core architecture: Folklore-inspired residual layers.} Our GNN's main building block is a residual layer that processes hidden states for all node pairs $(h^{t}_{i\to j})_{i,j}\in \mathbb{R}^{n\times n\times d}$, producing updated states $(h^{t+1}_{i\to j})_{i,j}\in \mathbb{R}^{n\times n\times d}$:
\begin{align}\label{eq:layer}
    h^{t+1}_{i\to j} = h^{t}_{i\to j} + m_1\left( h^{t}_{i\to j}, \sum_\ell h^{t}_{i\to \ell} \odot m_0(h^{t}_{\ell\to j})\right),
\end{align}
where $m_0:\mathbb{R}^d \to \mathbb{R}^d$ and $m_1:\mathbb{R}^{2d} \to \mathbb{R}^d$ are multilayer perceptrons (MLPs) with graph normalization layers, and $\odot$ denotes component-wise multiplication.

This design differs from the original Folklore-type GNN \citep{maron2019provably} in three ways:
\begin{itemize}
    \item \textbf{Residual connections}: The skip connection $h^{t}_{i\to j} + (\cdot)$ is a faithful neuralization of the 2-FWL update rule, which carries the colour from step $t$ forward and aggregates new evidence on top of it. We initially adopted residuals for the usual training-stability reasons, but they turn out to also be the structurally correct choice from the 2-FWL perspective. Stable optimization is a side benefit, not the primary motivation.
    \item \textbf{Graph normalization}: Inspired by \cite{cai2021graphnorm}, this ensures well-behaved tensor magnitudes across different graph sizes.
    \item \textbf{Single MLP in the multiplicative branch}: PPGN \cite{maron2019provably} applies an MLP to \emph{both} arguments of the component-wise product to argue, via universal approximation of power-sum polynomials, that each layer is as expressive as one 2-FWL step. We apply $m_0$ on only one branch, reducing per-layer compute and parameter count at the cost of a strict layer-by-layer expressivity guarantee: a single FGNN layer is not as expressive as a single PPGN layer. We show in Section~\ref{sec:expressivity} (Theorem~\ref{thm:fgnn-ppgn}) that this is recovered with doubled depth: a $2K$-layer FGNN of width $4d$ exactly simulates any $K$-layer PPGN of width $d$, and FGNN therefore inherits the full 2-FWL expressivity of PPGN.
\end{itemize}

\textbf{Input and output transformations.} The complete architecture consists of three main components:

\begin{enumerate}
    \item \textbf{Input embedding}: The adjacency matrix $A \in \{0,1\}^{n \times n}$ is embedded into the initial hidden state $h^{0}_{i\to j} \in \mathbb{R}^{n\times n\times d}$ using a learned embedding layer that encodes edge presence/absence.

    \item \textbf{Residual processing}: Multiple residual layers \eqref{eq:layer} transform the node-pair representations, capturing complex structural relationships.

    \item \textbf{Node feature extraction}: The final tensor $h^{k}_{i\to j} \in \mathbb{R}^{n\times n\times d}$ is converted to node features $\mathbb{R}^{n\times d}$ via max-pooling over the second spatial dimension: $\text{node}_i = \max_j h^{k}_{i\to j}$.
\end{enumerate}

\textbf{Ranking integration for chained networks.} The networks $g^{(1)}, g^{(2)}, \ldots$ must incorporate ranking information in addition to graph structure. For a graph of size $n$, the node at position $k\in\{0,\dots,n-1\}$ of the ordering by increasing score is represented by its relative rank $\tilde r=k/n\in[0,1)$ (the implementation stores the increasing order, which is equivalent to the decreasing order of \eqref{eq:ord}). Only this relative value is used, never the absolute position, which is why a model trained on $n=500$ can be evaluated on larger graphs. Ties between equal scores are ordered by the sorting routine (\texttt{numpy.argsort}) applied to the score vector, without any use of labels or of the planted permutation. In a chained FGNN, the input of the pair tensor has two channels: the adjacency matrix, and a second channel whose diagonal entry $(j,j)$ carries the relative rank $\tilde r_j$ of node $j$ (zero elsewhere), so that rank information enters the pair state on the diagonal and spreads through the residual pair-message-passing layers. After max-pooling, the node feature is concatenated with a sinusoidal positional encoding of the same relative rank (followed by a small MLP and a normalization), and $S^{A\to B}$ is the dot product of these concatenated node features. The similarity therefore decomposes as a structural term plus a learned kernel of the two relative ranks, but the structural term itself depends on the full rank fields through the pair tensor, so the similarity is not merely an additive sum of a rank-independent structural score and a rank score.

This architecture provides the expressiveness needed to capture global graph properties while remaining trainable through the sequential training procedure described in Section \ref{sec:training_chaining}. While scalability remains a limitation for very large graphs, the architecture proves highly effective for the graph sizes considered in our experiments (up to 1000 nodes).

\subsection{Expressivity: FGNN matches PPGN up to depth doubling}\label{sec:expressivity}

The FGNN block of \eqref{eq:layer} applies a single MLP $m_0$ inside the bilinear pair-aggregation, in contrast with PPGN \citep{maron2019provably} which applies one MLP to \emph{each} argument. This raises the concern that FGNN is strictly less expressive than PPGN at the layer level. We prove this is an artefact of the layer-by-layer comparison: at the level of the formal architecture (entry-wise MLPs, no graph normalisation), \textbf{any $K$-layer PPGN can be exactly simulated by a $2K$-layer FGNN of tensor-channel width $4d$}, and FGNN therefore inherits the full 2-FWL expressivity of PPGN. 

\paragraph{MLP class.} Throughout this section, the MLP class $\mathcal{M}$ consists of feed-forward networks with arbitrary finite depth, arbitrary finite hidden widths, and a continuous activation $\sigma$ that satisfies the \emph{identity-realisability} property below.

\begin{definition}[Identity-realisable activation]
\label{def:identity-realisable}
An activation $\sigma : \mathbb{R} \to \mathbb{R}$ is identity-realisable if there exist constants $a, b, c \in \mathbb{R}$ such that $a\,\sigma(x) + b\,\sigma(-x) + c = x$ for all $x$.
\end{definition}

ReLU obviously satisfies $\sigma(x) - \sigma(-x) = x$.

\begin{lemma}[MLP closure]
\label{lem:mlp-closure}
Let $\sigma$ be identity-realisable and $\mathcal{M}$ the corresponding MLP class. Then $\mathcal{M}$ is closed under: (1) affine pre-composition $f \circ A$; (2) affine post-composition $B \circ f$ (in particular $-f \in \mathcal{M}$); (3) parallel branching $x \mapsto (f_1(x), f_2(x))$; (4) composition $f \circ g$; (5) the identity map $\mathbb{R}^p \to \mathbb{R}^p$ is in $\mathcal{M}$, with one hidden layer of width $2p$. The constructed MLPs have width bounded by the sum of those of the constituents plus an additive constant depending only on input/output dimensions and on $\sigma$. The depth is bounded by the maximum of those of the constituents for operations (1)--(3) and (5), and by the sum of their depths for operation (4).
\end{lemma}
\begin{proof}
(1)--(2) absorb $A,B$ into the first/last affine layer of $f$. (3) places $f_1, f_2$ in parallel branches and concatenates their outputs (padding the shallower branch with identity gadgets to match depths). (4) stacks the layers. (5) follows from identity-realisability: write $x_i = a\sigma(x_i) + b\sigma(-x_i) + c$ component-wise, using $2p$ hidden units.
\end{proof}

\begin{definition}[PPGN block]
\label{def:ppgn-app}
A PPGN block with parameters $(M_1, M_2, M_3)$, where $M_1, M_2 : \mathbb{R}^d \to \mathbb{R}^d$ and $M_3 : \mathbb{R}^{2d} \to \mathbb{R}^d$ are MLPs in $\mathcal{M}$ applied entry-wise, maps $h \in \mathbb{R}^{n \times n \times d}$ to
\[
\mathrm{PPGN}(h)_{ij} \;=\; M_3\!\left( h_{ij},\; \sum_{\ell=1}^{n} M_1(h_{i\ell}) \odot M_2(h_{\ell j}) \right).
\]
\end{definition}

The FGNN block is the residual block of \eqref{eq:layer}, with $m_0, m_1 \in \mathcal{M}$ applied entry-wise and \emph{without} graph normalisation; we denote it $\mathrm{FGNN}(h)$. Graph normalisation is treated separately in Remark~\ref{rem:graphnorm-app}.

\begin{theorem}[FGNN simulates PPGN with depth doubling]
\label{thm:fgnn-ppgn}
Let $K \in \mathbb{N}$ and let $\mathcal{P} = \mathrm{PPGN}_K \circ \cdots \circ \mathrm{PPGN}_1$ be a $K$-layer PPGN of channel dimension $d$ with MLP parameters $(M_1^{(k)}, M_2^{(k)}, M_3^{(k)})$ for $k = 1, \ldots, K$. There exist FGNN blocks $\mathrm{FGNN}_1, \ldots, \mathrm{FGNN}_{2K}$ of channel dimension $d' = 4d$ such that, for every input $h \in \mathbb{R}^{n \times n \times d}$,
\[
\iota^{-1}\!\left( \mathrm{FGNN}_{2K} \circ \cdots \circ \mathrm{FGNN}_1\!\left(\iota(h)\right) \right) \;=\; \mathcal{P}(h),
\]
where $\iota : \mathbb{R}^d \hookrightarrow \mathbb{R}^{4d}$ embeds into the first channel block ($\iota(x) = (x, 0, 0, 0)$) and $\iota^{-1}$ extracts it. Moreover, the internal hidden widths of the FGNN MLPs are bounded above by a constant times $W^\star + d$, where $W^\star$ is the maximum hidden width of $M_1^{(k)}, M_2^{(k)}, M_3^{(k)}$ over all $k$, the constant depending only on $\sigma$. The additive $d$ term reflects the identity gadgets on $\mathbb{R}^d$ used in the construction (each of width $2d$ by Lemma~\ref{lem:mlp-closure}(5)).
\end{theorem}

\begin{proof}
It suffices to simulate one PPGN block: the construction below maps $\iota$-form inputs to $\iota$-form outputs and therefore composes cleanly across layers ($K = 0$ is vacuous). Write every $x \in \mathbb{R}^{4d}$ as $x = (x^{(1)}, x^{(2)}, x^{(3)}, x^{(4)})$ with each block in $\mathbb{R}^d$.

For a PPGN block with parameters $(M_1, M_2, M_3)$, define the first FGNN block by
\[
m_0^{(1)} \equiv 0, \qquad m_1^{(1)}(x, y) = \bigl(0,\, M_1(x^{(1)}),\, M_2(x^{(1)}),\, 0\bigr).
\]
By Lemma~\ref{lem:mlp-closure}, $m_1^{(1)} \in \mathcal{M}$ (drop $y$ via affine pre-composition; run $M_1, M_2$ in parallel; place in blocks $c_2, c_3$ via affine post-composition). The first residual yields
\[
h^1_{ij} = \iota(h)_{ij} + m_1^{(1)}(\iota(h)_{ij}, 0) = \bigl( h_{ij},\, M_1(h_{ij}),\, M_2(h_{ij}),\, 0 \bigr).
\]
For the second FGNN block, set $m_0^{(2)}(z) = (0, z^{(3)}, 0, 0)$ (an affine map; in $\mathcal{M}$ via Lemma~\ref{lem:mlp-closure}(5)+(2)). The bilinear sum then evaluates to
\begin{align*}
\sum_\ell h^1_{i\ell} \odot m_0^{(2)}(h^1_{\ell j})
&= \sum_\ell \bigl( h_{i\ell}, M_1(h_{i\ell}), M_2(h_{i\ell}), 0 \bigr) \odot \bigl( 0, M_2(h_{\ell j}), 0, 0 \bigr) \\
&= \Bigl( 0,\; \underbrace{\textstyle\sum_\ell M_1(h_{i\ell}) \odot M_2(h_{\ell j})}_{=:\, S_{ij}},\; 0,\; 0 \Bigr).
\end{align*}
Define $m_1^{(2)}(x, y) = \bigl( M_3(x^{(1)}, y^{(2)}) - x^{(1)},\, -x^{(2)},\, -x^{(3)},\, 0 \bigr)$. By Lemma~\ref{lem:mlp-closure} (extract $x^{(1)}, y^{(2)}$ via affine pre-composition, compose with $M_3$, subtract identity gadget on $x^{(1)}$; negate identity gadgets on $x^{(2)}, x^{(3)}$; combine via parallel branching), $m_1^{(2)} \in \mathcal{M}$. Applying the second residual,
\[
h^2_{ij} = h^1_{ij} + m_1^{(2)}\bigl( h^1_{ij}, (0, S_{ij}, 0, 0) \bigr) = \bigl( M_3(h_{ij}, S_{ij}),\, 0,\, 0,\, 0 \bigr) = \iota\bigl( \mathrm{PPGN}(h)_{ij} \bigr),
\]
which is in $\iota$-form, so the cycle composes for the next PPGN block. The width claim follows from Lemma~\ref{lem:mlp-closure}: each constructed MLP uses parallel copies of $M_1, M_2, M_3$, affine projections, and a constant number of identity gadgets at $2d$ each.
\end{proof}

\begin{corollary}[2-FWL expressivity]
\label{cor:fgnn-2fwl}
Equip PPGN with any graph-level read-out $R$ (e.g.\ summation over the diagonal of the final-layer tensor followed by an MLP), and equip the simulating FGNN with the read-out $R\circ\iota^{-1}$ on the final $4d$-channel tensor. For any pair of graphs $(G_1, G_2)$ that the 2-FWL test distinguishes, there exists a finite-depth FGNN with this read-out that produces different graph-level representations on $G_1$ and $G_2$.
\end{corollary}
\begin{proof}
By \cite{maron2019provably}, a sufficiently deep PPGN with the chosen read-out distinguishes any 2-FWL-distinguishable pair. Theorem~\ref{thm:fgnn-ppgn} applied at that depth produces, after $\iota^{-1}$, the same per-pair tensor as the PPGN; the read-outs then coincide.
\end{proof}

\begin{remark}[Converse direction]
PPGN is at least as expressive as FGNN at the block level: setting $M_1 = \mathrm{id}_{\mathbb{R}^d}$ in Definition~\ref{def:ppgn-app} reproduces the FGNN bilinear term, and $M_3$ accommodates the residual connection. So FGNN $\equiv$ PPGN in expressivity, up to a constant factor in depth and tensor-channel width.
\end{remark}

\begin{remark}[Graph normalisation: scope of the result]
\label{rem:graphnorm-app}
The FGNN of our experiments composes each $m_0, m_1$ with a graph-normalisation operator $\mathrm{GN}$ that subtracts a per-channel mean over all $n^2$ pair entries and divides by a per-channel standard deviation. $\mathrm{GN}$ is a tensor-wide reduction; it is not entry-wise and therefore not in $\mathcal{M}$. The $\mathrm{GN}$-augmented FGNN is, strictly, outside the scope of the theorem above.

We do \emph{not} prove that adding $\mathrm{GN}$ preserves the equivalence: a naive attempt fails because after $\mathrm{GN}$ the channel-block $c_3$ of $h^1$ contains a per-channel rescaling of $M_2(h)$, not $M_2(h)$, and the bilinear identity producing $S_{ij}$ no longer holds verbatim. Whether some compensating construction recovers the equivalence with $\mathrm{GN}$ is left open. Theorem~\ref{thm:fgnn-ppgn} settles only the single-$m_0$ question --- which is the substantive concern; $\mathrm{GN}$ is a separate practical ingredient. Empirically, the $\mathrm{GN}$-augmented FGNN solves correlated regular graph alignment (Section~\ref{sec:results}), a regime where 1-WL provably fails, providing direct evidence that the trained network exhibits expressivity strictly above 1-WL.
\end{remark}

\subsection{Technical details for the GNN architecture and training}\label{sec:technical}

By default, we use MLPs for the functions $m_0$ and $m_1$ in \eqref{eq:layer} with 2 hidden layers of dimension $256$.
In all our experiments, we take a GNN with 2 residual layers. We used the Adam optimizer with a learning rate of $10^{-4}$ and the scheduler ReduceLROnPlateau from PyTorch with a patience parameter of 3.

For $\Proj$, we use the function {\bf linear\_sum\_assignment} from {\bf scipy.optimize} and for $\FAQ$, we use the function {\bf quadratic\_assignment} from the same library. SciPy is a set of open source (BSD licensed) scientific and numerical tools for Python. In order to compute $D_\cx$ solving \eqref{eq:convrelax}, we implemented the Frank-Wolfe algorithm.

For the training and inference, we used Nvidia RTX8000 48GB and Nvidia A100 80GB. For graphs of size 500, we train on 200 graphs and validate on 100 graphs for 100 epochs.
We run for $L=15$ steps of chaining (obtaining 16 trained FGNNs: $f, g^{(1)},\dots, g^{(15)}$), each stage being trained for $100$ epochs on the outputs of the previous stage; the reduced three-stage chain of Section~\ref{sec:compute} was trained in about two GPU-hours. The training noise levels of the released chains are $0.22$ (sparse), $0.24$ (dense) and $0.11$ (regular).
The PyTorch code, the pretrained chains and the per-pair results of all tables are available at \url{https://github.com/mlelarge/chaining-gnn-graph-alignment}; the degree-normalized and random-order ablations of Appendix~\ref{sec:ablations} are options of the released pipeline.

\subsection{Ranking ablations and per-stage analysis}\label{sec:ablations}

All experiments in this section use the sparse Erd\H{o}s--R\'enyi benchmark ($n=500$, $d=4$) and a $30$ test pairs per noise level (same as in Section~\ref{sec:previouser}); the reference chain is the released chain trained at $p_\noise=0.22$ with the raw score.

\paragraph{Protocol.} The \emph{degree-normalized} variant ranks node $i$ by $\score(i)/\max(\deg_A(i),1)$ instead of $\score(i)$. The \emph{random order} keeps the decoded matching but assigns the $n$ rank values to the matched pairs in a uniformly random order, drawn independently at every stage; matched nodes still share a rank value, so the matching remains recoverable from the two rank fields, but the values no longer indicate confidence. Each variant is evaluated in two ways: \emph{at inference}, the raw-trained chain is run with the modified ranking (a stress test, since the stages then receive an ordering convention they were not trained on); \emph{retrained}, a new chain is trained from scratch with the modified ranking at every stage, with the same architecture, data and chain length as the reference chain. Test accuracies with $\FAQ$ decoding are given in Table~\ref{tab:rank-ablation}.

\begin{table}[!h]
  \caption{Ranking ablations on sparse Erd\H{o}s--R\'enyi graphs ($n{=}500$, $d{=}4$): accuracy with $\FAQ$ decoding on $30$ fixed test pairs per noise level. \emph{Raw} is the released chain, trained at $p_\noise{=}0.22$ with the raw score. \emph{At inference}: the raw-trained chain run with the modified ranking. \emph{Retrained}: a chain trained from scratch with the modified ranking at every stage, with the same architecture, data and chain length.}
  \label{tab:rank-ablation}
\begin{center}
\begin{small}
\begin{sc}
\begin{tabular}{lccccc}
\toprule
 & & \multicolumn{2}{c}{degree-normalized score} & \multicolumn{2}{c}{random order} \\
\cmidrule(lr){3-4} \cmidrule(lr){5-6}
$p_\noise$ & raw (trained) & at inference & retrained & at inference & retrained \\
\midrule
0.15 & 0.951 & 0.941 & 0.952 & 0.914 & 0.947 \\
0.20 & 0.928 & 0.751 & 0.924 & 0.346 & 0.802 \\
0.25 & 0.884 & 0.272 & 0.840 & 0.092 & 0.331 \\
0.30 & 0.440 & 0.081 & 0.465 & 0.040 & 0.097 \\
0.35 & 0.033 & 0.023 & 0.033 & 0.017 & 0.027 \\
\bottomrule
\end{tabular}
\end{sc}
\end{small}
\end{center}
\vskip -0.2in
\end{table}

\paragraph{Training trajectories under the ablations.} Table~\ref{tab:link-traj} gives the validation accuracy (with $\Proj$ decoding, at the training noise) of each stage of the three retrained chains. The first stage receives no ranking and is identical for the three chains. With either score-based ranking, the accuracy rises steeply over the first stages and saturates; the degree-normalized chain tracks the raw chain within six points at every stage. The random-order chain also improves monotonically, which confirms that the transported matching alone carries usable signal, but it plateaus far below the score-ordered chains and is still climbing by about $0.01$ per stage at the end of the chain: the score order reaches by the third stage an accuracy that the random order does not reach after fifteen. This is the practical content of the ranking: the transported matching makes chaining work at all, and the confidence ordering makes it reach useful accuracy in a feasible number of stages.

\begin{table}[h]
\centering
\caption{Per-stage validation accuracy ($\Proj$ decoding, training noise $p_\noise=0.22$, sparse Erd\H{o}s--R\'enyi graphs) of chains trained with the raw score, the degree-normalized score and a random order. Stage $0$ is the network $f$, which receives no ranking.}
\label{tab:link-traj}
\begin{footnotesize}
\setlength{\tabcolsep}{3.5pt}
\begin{tabular}{l*{15}{c}}
\toprule
stage & 0 & 1 & 2 & 3 & 4 & 5 & 6 & 7 & 8 & 9 & 10 & 11 & 12 & 13 & 14 \\
\midrule
raw score & .09 & .18 & .30 & .43 & .55 & .66 & .73 & .77 & .80 & .82 & .83 & .84 & .85 & .86 & .87 \\
degree-normalized & .09 & .17 & .27 & .37 & .49 & .59 & .66 & .71 & .74 & .76 & .77 & .79 & .80 & .81 & .81 \\
random order & .09 & .12 & .13 & .15 & .16 & .18 & .18 & .19 & .20 & .21 & .23 & .23 & .25 & .26 & .27 \\
\bottomrule
\end{tabular}
\end{footnotesize}
\end{table}

\paragraph{Correction and retention.} To test directly whether later stages correct the errors of earlier ones rather than merely preserving good matches, we track every node of the $30$ test pairs at test noise $p_\noise=0.2$ through the released chain ($\Proj$ decoding at each stage). For each transition $k\to k{+}1$ we report the \emph{correction rate}, the probability that a node that is misassigned at stage $k$ and lies in the bottom half of stage $k$'s ranking is correctly assigned at stage $k{+}1$, and the \emph{retention rate}, the probability that a node correctly assigned at stage $k$ and lying in the top half of the ranking is still correct at stage $k{+}1$ (Table~\ref{tab:correction}). At its peak, one stage corrects $35\%$ of all bottom-ranked errors while retaining essentially every top-ranked correct match. The correction rate first rises, as the set of reliable matches grows, then falls as the remaining error pool becomes smaller and harder. This rise separates two readings of what the ranking does. If the ranking only broke ties among structurally indistinguishable nodes, or only sharpened the similarities of nodes that are already correctly matched, a wrong node at the bottom of the ranking would be corrected at a rate independent of how many reliable matches the previous stage provides. Instead, the probability that such a node is corrected grows with the number of reliable matches, from $0.14$ to $0.35$, which is what one expects if the reliable matches serve as evidence for re-solving the remaining nodes (Section~\ref{sec:chaining}).
 Consistently, the mean $\nce$ of the $\Proj$-decoded alignment over the $30$ pairs increases monotonically along the chain, $115\to226\to369\to522\to633\to701\to736\to756\to767\to771\to775\to777$, and then plateaus; see also Figure~\ref{fig:iter_sparse}.

\begin{table}[h]
\centering
\caption{Correction and retention rates per stage of the released sparse chain at test noise $p_\noise=0.2$ ($30$ test pairs). acc@$k$ is the $\Proj$ accuracy of stage $k$; counts in parentheses are corrected nodes over bottom-half misassigned nodes.}
\label{tab:correction}
\begin{small}
\begin{tabular}{lccc}
\toprule
stage $k\to k{+}1$ & acc@$k$ & correction (bottom half, wrong at $k$) & retention (top half, correct at $k$) \\
\midrule
$0\to1$ & 0.12 & 0.14 (1019/7123) & 0.92 \\
$1\to2$ & 0.23 & 0.25 (1698/6937) & 0.97 \\
$2\to3$ & 0.40 & 0.33 (2147/6506) & 0.99 \\
$3\to4$ & 0.58 & 0.35 (1883/5422) & 1.00 \\
$5\to6$ & 0.79 & 0.27 (815/2991) & 1.00 \\
$8\to9$ & 0.88 & 0.11 (182/1671) & 1.00 \\
\bottomrule
\end{tabular}
\end{small}
\end{table}

\begin{figure}[h]
  \centering
  \includegraphics[width=0.55\linewidth]{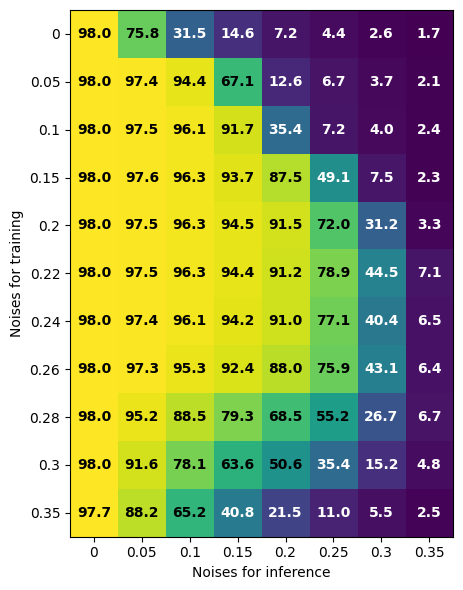}
  \caption{Each line corresponds to chained FGNNs trained at a given level of noise and evaluated across all different level of noises. Performances are $\acc$ (in \%) for sparse Erd\H{o}s-R\'enyi graphs with $\Proj$ as post-processing.}
  \label{fig:noise}
\end{figure}

\paragraph{Train/test mismatch and chain length.} Each chain is trained at a single noise level (sparse $0.22$, dense $0.24$, regular $0.11$) and evaluated on the full test grid (Tables~\ref{tab:ER_sparse_acc}--\ref{tab:Reg_nce}, Figure~\ref{fig:noise}). Performance is stable at all noise levels at or below the training noise and degrades gracefully beyond it: chaining helps once the first stages reach a basin with enough reliable pairs, and otherwise falls back toward single-stage behavior; in Table~\ref{tab:ER-Reg} it never falls below the single-FGNN baseline. Accuracy and $\nce$ saturate along the chain (Table~\ref{tab:accuracy_comparison}, Figure~\ref{fig:iter_sparse}), so beyond the plateau additional stages or loops cost computation but not accuracy; the number of stages is a compute budget rather than a sensitive hyperparameter, and the reduced three-stage chain of Section~\ref{sec:compute} already recovers most of the accuracy.

\paragraph{Why the gains are larger on sparse graphs.} The transition is sharper on dense graphs and the window where chaining helps is narrower. Two quantities explain this. In an Erd\H{o}s--R\'enyi graph with average degree $d$ and edge probability $\lambda=d/n$, the relative spread of degrees is $\sqrt{(1-\lambda)/d}\approx 1/\sqrt d$: large when sparse, so that heterogeneous degrees provide strong anchors, and shrinking as $d$ grows, until the graph resembles the regular case where the first assignment is degenerate. The evidence that a reliably matched neighbor pair provides about a candidate pair, the witness count of Section~\ref{sec:chaining}, behaves in the same way: it corroborates a true candidate with probability $\lambda(1-p_\noise)$ but a wrong one with probability $\lambda^2$, a contrast ratio $(1-p_\noise)/\lambda$ that shrinks as the graph densifies or as the noise grows. The advantage of chaining is therefore one of relative contrast, not of sparse instances being uniformly easier.

\paragraph{Uninformative rank fields.} The random-order ablation keeps the matching in the rank fields. A \emph{fully} random rank field, drawn independently for the two graphs, carries no information about the correspondence and cannot help in the following sense.
\begin{proposition}\label{prop:random-field}
Let the rank fields $(r^A,r^B)$ given to a stage be drawn independently of $(G_A,G_B,\pi^\star)$. For any loss function, the minimal expected loss over predictors that observe $(A,B,r^A,r^B)$ equals the minimal expected loss over predictors that observe $(A,B)$ only.
\end{proposition}
\begin{proof}
By independence, the conditional law of $\pi^\star$ given $(A,B,r^A,r^B)$ equals its conditional law given $(A,B)$. Hence a Bayes-optimal predictor for the enlarged observation can be chosen to depend on $(A,B)$ only, and the two Bayes risks coincide; any (possibly randomized) predictor of the enlarged observation has expected loss at least this Bayes risk.
\end{proof}
Three remarks delimit what the proposition says. It concerns the information available to an ideal predictor, not what a given finite GNN can compute: random inputs independent of the instance carry no information about $\pi^\star$, yet they can still help a GNN by breaking symmetries between nodes \citep{sato2021random,abboud2021surprising}. It does not cover the keep-best rule, which selects among several stages by their $\nce$ and is a separate source of gains. And it does not apply to the random order of the ablation, which gives matched nodes the same rank value and therefore carries the matching, nor to the score order.

\subsection{More related work}\label{sec:related}

Supervised learning approaches for the graph matching problem have been extensively studied in the computer vision literature \citep{wang2021neural}, \citep{rolinek2020deep}, \citep{zanfir2018deep}, \citep{gao2021deep}, \citep{yu2021deep}, \citep{jiang2022glmnet}. \citep{fey2020deep} is closely related to our work and proposes a two-stage architecture similar to our chaining procedure with MPNNs. The first stage is the same as our first step but with an MPNN instead of our FGNN. Then the authors propose a differentiable, iterative refinement strategy to reach a consensus of matched nodes. All these works assume that non-topological node features are available and informative. This is a setting favorable to GNNs as node-based GNNs are effective in learning how to extract useful node representations from high-quality non-topological node features. In contrast, we focus on the pure combinatorial problem where no side information is available.
In \citep{li2019graph}, graph matching networks take a pair of graphs as input and compute a similarity score between them. This algorithm can be used to compute the value of the graph matching \eqref{eq:gm} but does not give the optimal permutation $\pi^{A\to B}$ between the two graphs which is the main focus of our work.
\citep{lagesse2025graph} propose a benchmarking methodology for GNNs based on graph alignment, and show that the same task also serves as an effective objective for unsupervised GNN pre-training.

Regarding benchmarks for the combinatorial GAP, we are not aware of any publicly available dataset. The GAP can be seen as a particular version of the QAP and some algorithms designed for the GAP can be used for QAP instances (i.e. with weighted adjacency matrices). This is the case for the convex and indefinite relaxations presented in Section \ref{sec:cont} which can be used with real-valued matrices. In particular, \citep{lyzinski2015graph} shows very good performances of $\FAQ$ on some QAP instances from \citep{burkard1997qaplib}. These instances are small (from 12 to 40 nodes) with full (integer-valued) matrices. They are very far from the distribution of correlated random graphs used for training in our work and we do not expect good performances for such out-of-distribution instances for any supervised learning algorithm.

\subsection{Model-based versus simulation-based algorithms}\label{sec:theo}

As explained in Section \ref{sec:datasets}, we train and evaluate our supervised learning algorithms on correlated random graphs. This choice connects our work to a rich theoretical literature on the correlated Erd\H{o}s-R\'enyi random graph ensemble, which has been extensively studied from an information-theoretic perspective.

\subsubsection{Theoretical Foundations and Limits}

The theoretical analysis begins with \cite{10.1145/2964791.2901460}, which establishes the information-theoretic limit for exact recovery of $\pi^{\star}$ as the number of nodes $n$ tends to infinity. In the sparse regime, where the average degree $d$ remains constant as $n\to\infty$, exact recovery becomes impossible. Subsequent work by \cite{ganassali2021impossibility} and \cite{ding2023matching} demonstrates that partial recovery of $\pi^{\star}$ is only possible when $p_\noise < 1-d^{-1}$.

For the correlated Erd\H{o}s-R\'enyi ensemble, the joint probability distribution is given by:
\begin{eqnarray*}
\Pb(G_A, G_B) \propto \left( \frac{(1-p_{\noise})(n-d(1+p_{\noise}))}{d p_{\noise}^2}\right)^{e(G_A \wedge G_B)},
\end{eqnarray*}
where $e(G_A \wedge G_B) = \sum_{i<j} A_{ij}B_{ij}$ counts the common edges between graphs $G_A$ and $G_B$. This distribution reveals a crucial insight: \textbf{the maximum a posteriori estimator of $\pi^*$ given $G_A$ and the permuted graph $G'_B$ is exactly a solution of the GAP on the $(G_A, G'_B)$ instance}.

\subsubsection{Model-based Approaches: Achievements and Limitations}

Recent theoretical advances have produced efficient polynomial-time algorithms \citep{ding2021efficient, fan2023spectral, ding2023polynomial, ganassali2024correlation, piccioli2022aligning} that approximate the probability distribution by exploiting structural properties like the local tree-like nature of sparse random graphs. These algorithms achieve partial recovery (positive accuracy) when $p_\noise$ is sufficiently small, though well below the information-theoretic threshold of $1-d^{-1}$.

However, a fundamental \textbf{algorithmic threshold} appears to exist. Recent work \citep{mao2023random, ganassali2024statistical} suggests that no efficient algorithm can succeed for $p_\noise > p_\algo = 1-\sqrt{\alpha} \approx 0.419$, where $\alpha$ is Otter's constant, even when the average degree $d$ is large.

While these model-based algorithms provide theoretical guarantees for correlated Erd\H{o}s-R\'enyi graphs, they suffer from significant practical limitations:
\begin{itemize}
\item \textbf{Narrow applicability}: Designed specifically for the correlated Erd\H{o}s-R\'enyi model with no guarantees outside this distribution
\item \textbf{Computational complexity}: Despite polynomial-time guarantees, running times are often impractical for real applications
\item \textbf{Limited scalability}: Most implementations prioritize mathematical rigor over computational efficiency
\end{itemize}

\cite{muratori2024faster} represents a notable exception, focusing on making message-passing algorithms \citep{ganassali2024correlation, piccioli2022aligning} more scalable while maintaining theoretical guarantees.

\subsubsection{FAQ: An Empirical Surprise}
\begin{figure}[ht]
  \begin{center}
  \centerline{\includegraphics[width=0.85\textwidth]{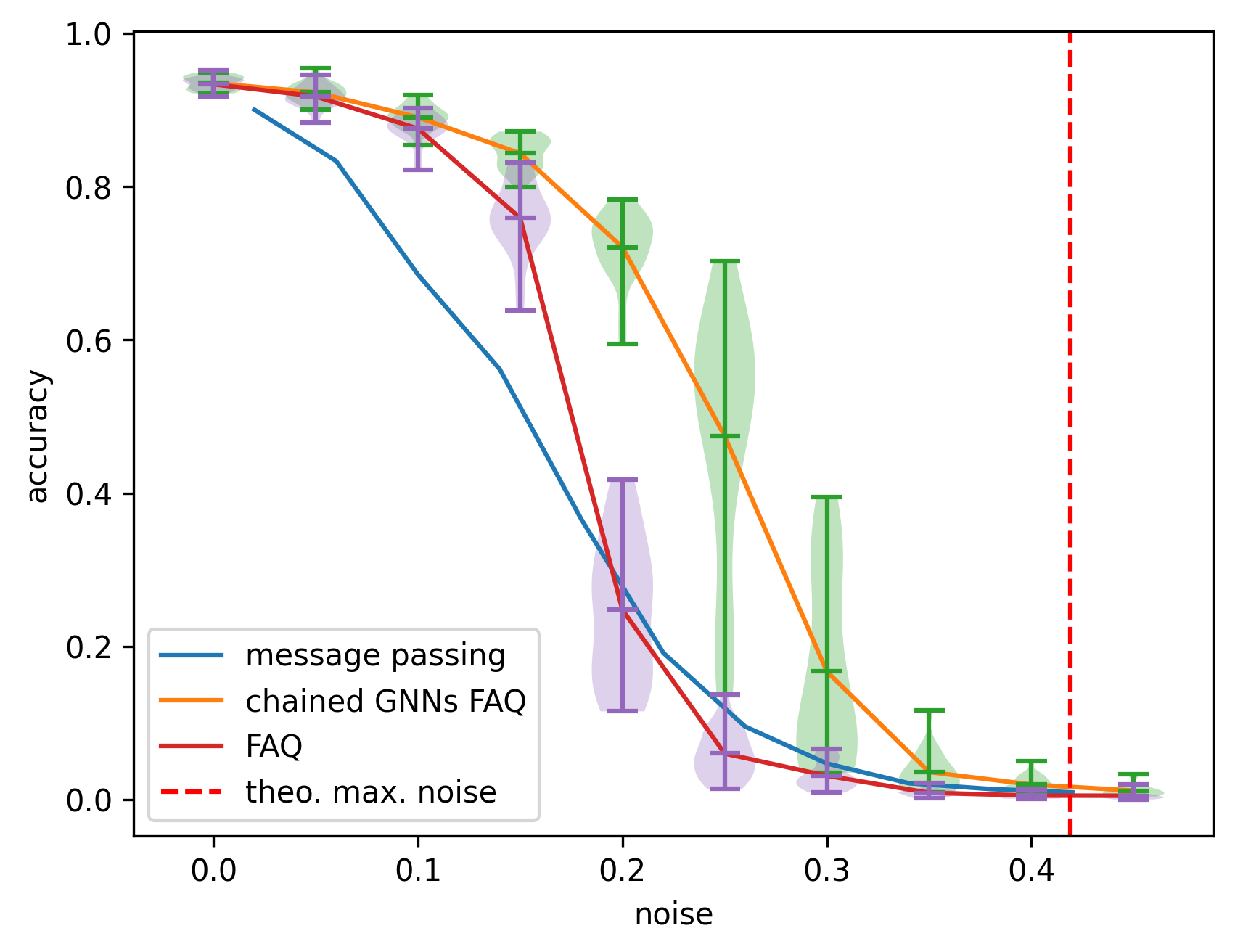}}
  \caption{Accuracy $\acc$ as a function of the noise level for correlated Erd\H{o}s-R\'enyi random graphs with size $n=1000$ and average degree $d=3$. Chained GNNs were trained at noise level $0.25$ and $\FAQ$ is used as the last step for the inference. The red curve labeled FAQ corresponds to $\FAQ(D_\cx)$ and the blue curve labeled message passing are results from \citep{muratori2024faster}. The dashed vertical line corresponds to the theoretical $p_\algo=1-\sqrt{\alpha}$ above which no efficient algorithm is known to succeed.}
  \label{fig:faq_message}
  \end{center}
\end{figure}
Remarkably, the $\FAQ$ algorithm---which was not designed specifically for any random graph model---empirically encounters the same algorithmic barrier predicted by theory. As shown in Figure \ref{fig:faq_message}, $\FAQ$'s performance degrades sharply near $p_\algo$, matching the theoretical predictions despite lacking formal guarantees for this setting. $\FAQ$ only underperforms compared to specialized message-passing methods \cite{muratori2024faster} when $p_\noise$ approaches $p_\algo$.

This empirical observation suggests that $\FAQ$, through its continuous relaxation approach, implicitly captures fundamental structural properties of the graph alignment problem that transcend specific random graph models.

\subsubsection{Our Simulation-based Approach}

To circumvent the computational challenges of maximizing the exact posterior (which, as shown above, corresponds exactly to solving the GAP), we adopt a \textbf{simulation-based approach}. Rather than deriving model-specific algorithms, we:
\begin{enumerate}
\item \textbf{Sample training data}: Generate pairs of graphs $(G_A, G_B)$ with known alignment permutations $\pi^{\star}$
\item \textbf{Learn mappings}: Train neural networks to map graph pairs to similarity matrices $S^{A\to B} \in \mathbb{R}^{n \times n}$
\item \textbf{Extract solutions}: Convert similarity matrices to permutations via projection or as $\FAQ$ initialization
\end{enumerate}

\paragraph{Key Advantages.} Our simulation-based approach offers several advantages over both model-based methods and traditional relaxations:

\textbf{Supervised learning with ground truth}: Unlike the convex relaxation \eqref{eq:convrelax}, we have access to ground truth permutations during training, enabling more informative loss functions.

\textbf{Better optimization objective}: Instead of the Frobenius norm used in convex relaxation, we employ cross-entropy loss, which provides more informative gradients for discrete matching problems.

\textbf{Generalization potential}: While trained on specific distributions, our learned representations may capture general structural patterns applicable beyond the training distribution.

\textbf{Hybrid capability}: Our similarity matrices can initialize traditional solvers like $\FAQ$, combining the benefits of learning and optimization approaches.

This simulation-based methodology bridges the gap between theoretical guarantees and practical performance, achieving strong empirical results while maintaining computational tractability.

\subsection{Relating GAP to Gromov-Hausdorff, Gromov-Monge and Gromov-Wasserstein distances for finite metric spaces}

We consider a simple case of discrete spaces with the same number of elements $n$ and where $A, B\in \R^{n\times n}$ are the distance matrices of two finite metric spaces $(X,d_X)$ and $(Y,d_Y)$, i.e. $A_{ij}=d_X(x_i,x_j)$ and $B_{ij}=d_Y(y_i,y_j)$. Recall that we denote by $\Scal_n$ the set of permutation matrices and by $\Dcal_n$ the set of doubly stochastic matrices. We also denote by $\Rcal_n$ the set of matrices $R\in \{0,1\}^{n\times n}$ such that $\sum_i R_{ij}\geq 1$ and $\sum_j R_{ij}\geq 1$.

The {\bf Gromov-Hausdorff distance} for finite metric spaces can be written as:
\begin{align}
 \label{eq:GH}   \GH_L(A,B) = \min_{R\in \Rcal_n}\max_{i,j,k,\ell}L(A_{ik}, B_{j\ell})R_{ij}R_{k\ell}
\end{align}
where $L(a,b)\geq 0$.
It is often desirable to smooth the max operator in \eqref{eq:GH} to a sum. This can be done by considering the related problem:
\begin{align}
\label{eq:GM} \GM_L(A,B) = \min_{R\in \Rcal_n}\sum_{i,j,k,\ell}L(A_{ik}, B_{j\ell}) R_{ij}R_{k\ell}
\end{align}
Note that if we replace the constraint $R\in \Rcal_n$ by $P\in \Scal_n$, we get the permutation-constrained version
\begin{align*}
    \GM_L^{\Scal}(A,B) = \min_{P\in \Scal_n}\sum_{i,j,k,\ell}L(A_{ik}, B_{j\ell}) P_{ij}P_{k\ell}
    = \min_{\pi\in \Scal_n}\sum_{i,j}L\left(A_{ij},B_{\pi(i)\pi(j)}\right),
\end{align*}
which is called {\bf Gromov-Monge distance}.

The {\bf Gromov-Wasserstein distance} is a relaxation of the Gromov-Hausdorff distance and is defined in \cite{memoli2011gromov}:
\begin{align}
\GW_L(A,B,p,q) = \min_{T\in \Ccal_{p,q}}\sum_{i,j,k,\ell} L(A_{ik}, B_{j\ell})T_{ij}T_{k\ell},
\end{align}
where $p,q$ are probability distributions on $X$, $Y$ and the minimum is taken over $\Ccal_{p,q} =\{T\in \R_+^{n\times n},\: T\bone = p, \: T^T\bone =q \}$.
Taking $p=q=\bone/n$ the uniform distribution, we have $\Ccal_{p,q}=\frac{1}{n}\Dcal_n$ and $\GW_L(A,B,\bone/n,\bone/n)$ is a relaxed version of $\GM_L(A,B)$.
We typically consider $L(a,b)= |a-b|^2$, and then we get:
\begin{align*}
    \GM^{\Scal}_{L^2}(A,B) = \min_{\pi\in \Scal_n}\sum_{i,j} (A_{ij}- B_{\pi(i)\pi(j)})^2,
\end{align*}
and with the simplified notation $\GW_{L^2}(A,B)=\GW_{L^2}(A,B,\bone/n,\bone/n)$,
\begin{align*}
n^2 \GW_{L^2}(A,B) &= \min_{D\in \Dcal_n}\sum_{i,j,k,\ell} (A_{ik}- B_{j\ell})^2D_{ij}D_{k\ell}\\
&= \min_{D\in \Dcal_n}\sum_{i,k}A^2_{ik}+\sum_{j\ell}B^2_{j\ell}-2\sum_{i,j,k,\ell}A_{ik} B_{j\ell}D_{ij}D_{k\ell}\\
&= \|A\|_F^2+\|B\|_F^2-2\max_{D\in \Dcal_n}\langle AD, DB \rangle.
\end{align*}

In the particular case where $A$ and $B$ are positive semidefinite matrices, i.e. $A=U^TU$ and $B=V^TV$, we have: $\langle AD, DB \rangle = \|UDV^T\|_F^2$ which is a convex function of $D$ and is always maximized at an extremal point of its constraint polytope $\Dcal_n$.
By Birkhoff's theorem, the extremal points of $\Dcal_n$ are permutation matrices. Therefore, we have: $n^2\GW_{L^2}(A,B) = \GM^{\Scal}_{L^2}(A,B)$ in this case. \cite{maron2018probably} shows that a similar result holds for Euclidean distances, when $A_{ij}=\|x_i-x_j\|_2$ and $B_{ij}=\|y_i-y_j\|_2$. Hence, we have:

\begin{proposition} \label{prop:OT}
  For $A$, $B$ Euclidean distance matrices, the indefinite relaxation \eqref{eq:indef} is tight and solves the GAP \eqref{eq:gm}. In this case, the GAP computes the Gromov-Monge distance and the indefinite relaxation computes $n^2$ times the Gromov-Wasserstein distance under uniform marginals.
\end{proposition}

\subsection{Notations used in tables}
\begin{itemize}
  \item $\acc$  $\FAQ(D_\cx)$ means the accuracy of $\FAQ$ algorithm initialized with $D_{cx}$.
  \item $\acc$  ChFGNN $\Proj$ means the accuracy of our chained FGNNs with $\Proj$ as the last step.
  \item $\acc$  ChFGNN $\FAQ$ means the accuracy of our chained FGNNs with $\FAQ$ as the last step.
  \item $\nce$  $\FAQ(D_\cx)$ means the number of common edges found by $\FAQ$ algorithm initialized with $D_{cx}$.
  \item $\nce$  $\FAQ(\pi^\star)$ means the number of common edges found by $\FAQ$ algorithm initialized with $\pi^\star$.
  \item $\nce$  ChFGNN $\Proj$ means the number of common edges found by our chained FGNNs with $\Proj$ as the last step.
  \item $\nce$  ChFGNN $\FAQ$ means the number of common edges found by our chained FGNNs with $\FAQ$ as the last step.
  \end{itemize}

 \subsection{Bernoulli graphs: Generalization properties for chained GNNs}

 \subsubsection{Training chained GNNs}\label{sec:training}

   \begin{figure}
       \centering
       \begin{minipage}{.5\textwidth}
         \centering
         \includegraphics[width=.9\linewidth]{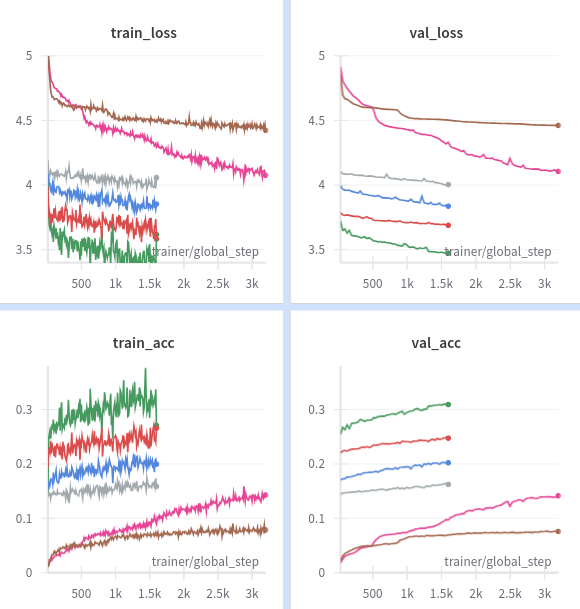}

       \end{minipage}%
       \begin{minipage}{.5\textwidth}
         \centering
         \centerline{\includegraphics[width=0.65\textwidth]{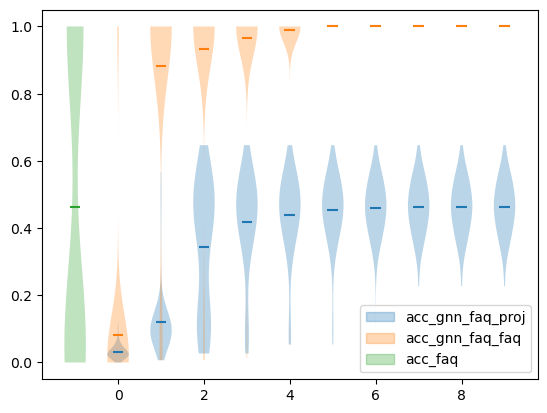}}
     \centerline{\includegraphics[width=0.66\textwidth]{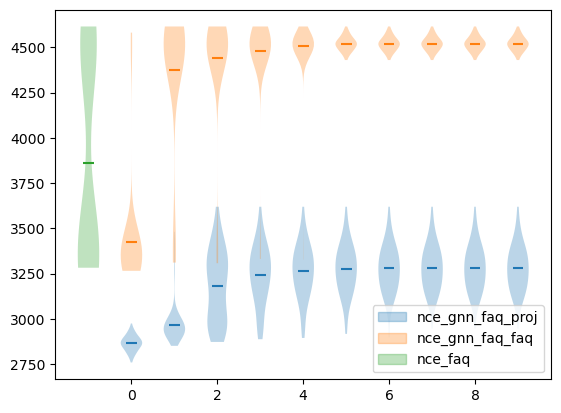}}
       \end{minipage}
       \caption{Left: Training chained GNNs. Each color corresponds to a different training and GNN: the first training (brown) reaches an accuracy below $0.1$. The second training (magenta) uses as input the output of the first training and get an accuracy $\approx 0.15$. The remaining trainings using the output of the previous training as input and reach higher and higher accuracy.\\
       Right top: $\acc$ bottom: $\nce$. First Violin plot (${\bf \_faq}$ green) for $\FAQ$, then all other Violin plots correspond to a different number of iterations $N_{\max}=0,1,\dots, 9$ of the chaining procedure (${\bf \_gnn\_faq\_proj}$ blue with $\Proj$ and ${\bf \_gnn\_faq\_faq}$ orange with $\FAQ$).}
       \label{fig:training_inference}
       \end{figure}

 For the same dataset as in \cite{lyzinski2015graph} (see Bernoulli graphs in Section \ref{sec:def_graphs}), we plot in Figure \ref{fig:training_inference} the training curves for the chained GNNs for the loss and the accuracy on the training set and the validation set.
 We do not see any overfitting here as values are similar on both sides.
 Chaining is very effective in this case: while the first training (brown) corresponding to the mapping $f$ in \eqref{eq:S0} saturates at an accuracy below $0.1$, the second training (magenta) corresponding to the mapping $g^{(1)}$ in \eqref{eq:S1} reaches a much higher accuracy because it uses the information about the graph matching contained in the output of the first training $f$. The curves for the remaining trainings are indeed ordered. This is due to the fact that for the training of $g^{(k+1)}$, we initialized it with the weights obtained after the training of $g^{(k)}$ in order to speed up the training. Since we observe a saturation in the learning of the $g^{(k)}$ for $k\geq 2$, we stop the training after half the number of epochs used for $f$ and $g^{(1)}$.

 \subsubsection{Efficient inference for chained FGNNs}\label{sec:inference}

 These results suggest an extreme form of looping: since $g^{(1)}$ allows to improve the accuracy of an initial guess (given by $f$),
 we can keep only the GNNs $f$ and $g^{(1)}$, and we loop through $g^{(1)}$ for a fixed number of steps $N_{\max}$.
 Figure \ref{fig:training_inference} gives the accuracy $\acc$ and the number of common edges $\nce$ defined in \eqref{eq:defmetrics} for the inference procedure as a function of the number of iterations $N_{\max}$ made on $g^{(1)}$. We give (in blue) the performances of $\Proj$, and (in orange) the performances of $\FAQ$ applied on the similarity matrix obtained after $L$ loops. We see that the Frank-Wolfe algorithm used in $\FAQ$ used as the last step of our chaining procedure is crucial to get better performances. Indeed, we need only $N_{\max}=5$ loops in order to get a perfect accuracy $\acc=1$ with $\FAQ$.

 We also give (in green) the performance of $\FAQ$ applied on the matrix $D_\cx$ (the default option in the $\FAQ$ algorithm). Indeed, $\FAQ(D_\cx)$ is able to find the correct permutation for the graph matching in $13\%$ of the cases and is stuck in a local maxima with a very small (less than $20\%$) accuracy otherwise. This bimodal behavior is due to the fact that $D_\cx$ gives very little information about the correct permutation. In contrast,
 {\bf the chaining procedure was able to learn a much better initialization than $D_{\cx}$ for $\FAQ$ allowing to improve the accuracy from $50\%$ to an exact accuracy.}
 \begin{table}[!b]
  \centering
  \begin{tabular}{lllllllll}
    \toprule
      & noise & 0.4 & 0.45 & 0.5 & 0.55 & 0.6 & 0.65 & 0.7 \\
    \midrule
     & $\acc$ $\Proj(D_\cx)$& 0.3428 & 0.1956 & 0.1209 & 0.0815 & 0.0552 & 0.0411 & 0.0309 \\
     & $\acc$ $\FAQ(D_\cx)$& 1.0 & 0.9954 & 0.9531 & 0.6910 & 0.2621 & 0.0959 & 0.0225 \\
     & $\nce$ $\Proj(D_\cx)$ & 3147.7 & 3000.0 & 2960.2 & 2945.8 & 2942.0 & 2942.4 & 2933.6 \\
      & $\nce$ $\FAQ(D_\cx)$& 4737.8 & 4622.8 & 4462.2 & 4056.1 & 3564.9 & 3408.0 & 3352.8 \\
    \midrule
    training 0.5 & $\acc$ ChFGNN $\Proj$& 0.9994 & 0.9962 & 0.9639 & 0.7842 & 0.3400 & 0.1442 & 0.0737 \\
      & $\acc$ ChFGNN $\FAQ$ & 1.0 & 1.0 & 1.0 & 0.9915 & 0.8949 & 0.5105 & 0.1267 \\
      & $\nce$ ChFGNN $\Proj$ & 4736.1 & 4617.1 & 4439.4 & 4025.8 & 3319.0 & 3085.4 & 3038.0 \\
      & $\nce$ ChFGNN $\FAQ$ & 4737.8 & 4629.0 & 4520.0 & 4395.8 & 4188.4 & 3747.2 & 3413.5 \\
    \bottomrule
    \end{tabular}
      \caption{Accuracy $\acc$ and number of common edges $\nce$ for Bernoulli graphs as a function of the noise $p_\noise$.}
      \label{tab:Bern}
\end{table}

 \begin{figure}[!t]
  \begin{center}
  \centerline{\includegraphics[width=0.35\textwidth]{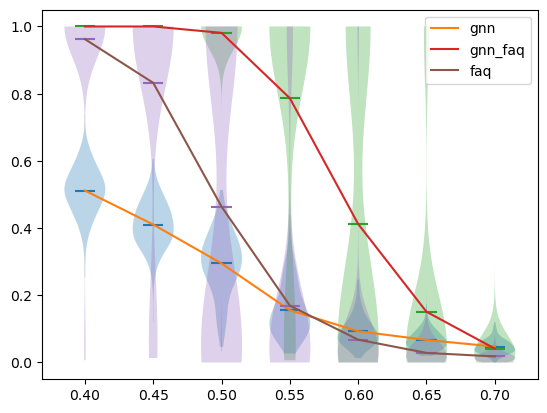}}
  \centerline{\includegraphics[width=0.36\textwidth]{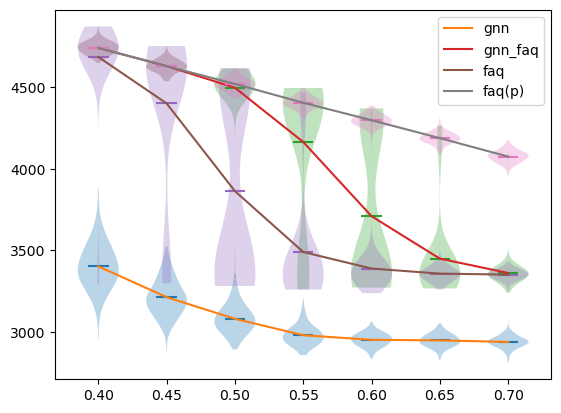}}
  \caption{Bernoulli graphs: $\acc$ (top) and $\nce$ (bottom) as a function of the noise level. Chained FGNNs were trained at noise level $0.5$. {\bf gnn} (resp.\ {\bf gnn\_faq}) for chained FGNNs with $\Proj$ (resp. $\FAQ$) as the last step. {\bf faq} for $\FAQ(D_\cx)$ and {\bf faq(p)} for $\FAQ(\pi^\star)$.}
  \label{fig:generalization}
  \end{center}
\end{figure}

We now explore the generalization properties of the chaining procedure by applying the inference procedure described in Section \ref{sec:inference} on datasets with different noise levels. The level of noise used during training (described in Section \ref{sec:training}) is $0.5$. Figure \ref{fig:generalization} gives the accuracy $\acc$ and the number of common edges $\nce$ for the inference procedure as a function of the noise level. We stop the inference loop when the $\nce$ obtained after applying $\FAQ$ to the similarity matrix is not increasing anymore. The red curve gives the performances of our chaining procedure with $\FAQ$ as the last step, the orange curve gives the performances of our chaining procedure with $\Proj$ as the last step. We compare our chaining procedure to $\FAQ(D_\cx)$ in brown and to $\FAQ(\pi^\star)$ in grey which corresponds to the maximum number of common edges for these noise levels. The curve for the accuracy of $\FAQ(D_\cx)$ is similar to the one obtained in \cite{lyzinski2015graph}. Our chaining procedure is able to generalize to noise levels different from the one used during training and outperforms $\FAQ(D_\cx)$ in all cases. Indeed with a noise level less than $0.5$, our chaining procedure recovers the correct permutation for the graph matching problem. Note that we did not try to optimize the performances of our chaining procedure with $\Proj$ as the last step, and they are indeed increasing if we allow for more loops.

\subsection{Additional results for sparse Erd\H{o}s-R\'enyi graphs}

\begin{table}[h]
  \caption{Accuracy ($\acc$) for sparse Erd\H{o}s-R\'enyi graphs as a function of the number ($L+1$) of trained FGNNs. Parentheses report the gain from extending inference to $N_{\text{loop}}=60$ loops instead of stopping after the $L+1$ trained chain steps. Last line: number of loops for chained FGNNs as a function of the noise $p_\noise$ to get optimal $\nce$ (with a maximal $N_{\text{loop}}=100$).}
  \label{tab:accuracy_comparison}
  \begin{center}
    \begin{small}
      \begin{sc}
        \begin{tabular}{llllll}
          \toprule
          Noise & 0.15 & 0.2 & 0.25 & 0.3 & 0.35 \\
          \midrule
L+1=2 & 0.28 (+0.02) & 0.15 (+0.02) & 0.08 (+0.01) & 0.04 (+0.01) & 0.02 (+0.00) \\
L+1=6 & 0.59 (+0.01) & 0.43 (+0.06) & 0.21 (+0.11) & 0.07 (+0.05) & 0.03 (+0.01) \\
L+1=10 & 0.85 (+0.01) & 0.72 (+0.06) & 0.43 (+0.13) & 0.11 (+0.19) & 0.04 (+0.03) \\
L+1=14 & 0.91 (+0.00) & 0.86 (+0.01) & 0.57 (+0.13) & 0.16 (+0.21) & 0.04 (+0.04) \\
L+1=16 & 0.92 (+0.00) & 0.88 (+0.01) & 0.61 (+0.12) & 0.19 (+0.26) & 0.04 (+0.04) \\
\midrule
\#loop &  15	& 23	& 88	& 91	& 73 \\
\bottomrule
          \end{tabular}
\end{sc}
\end{small}
\end{center}
\end{table}

\begin{figure}[ht]
  \begin{center}
  \includegraphics[width=0.45\textwidth]{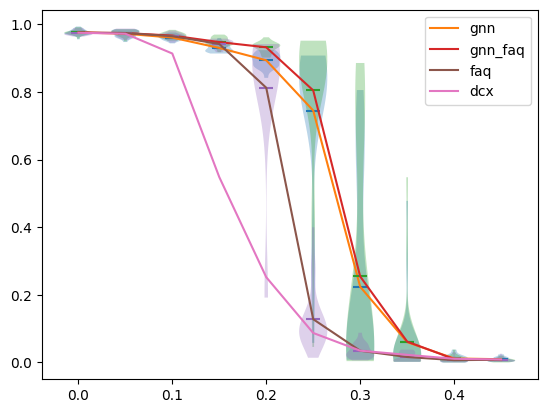}
  \includegraphics[width=0.46\textwidth]{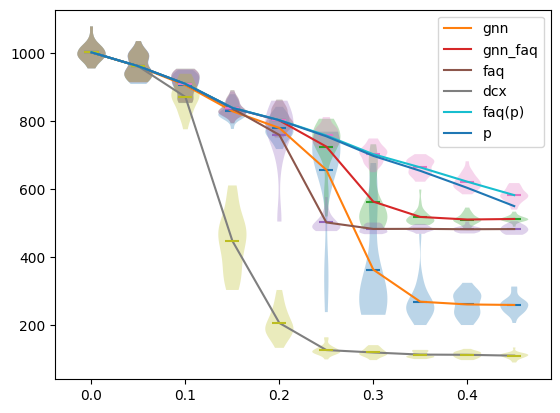}
  \caption{Sparse Erd\H{o}s-R\'enyi graphs: $\acc$ (top) and $\nce$ (bottom) as a function of the noise level. Chained FGNNs were trained at noise level $0.25$. {\bf gnn} (resp.\ {\bf gnn\_faq}) for chained FGNNs with $\Proj$ (resp. $\FAQ$) as the last step. {\bf faq} for $\FAQ(D_\cx)$ and {\bf faq(p)} for $\FAQ(\pi^\star)$.}
  \label{fig:sparse}
  \end{center}
\end{figure}

\begin{figure}[ht]
  \begin{center}
  \includegraphics[width=0.45\textwidth]{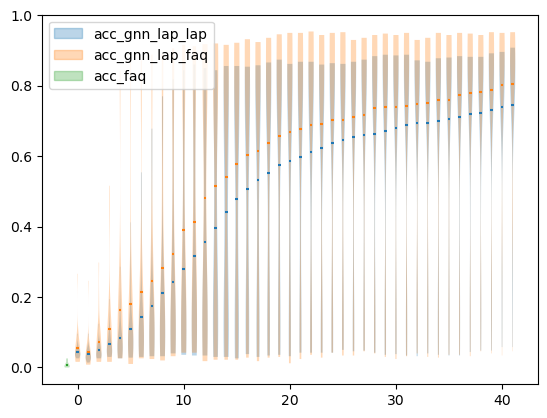}
  \includegraphics[width=0.46\textwidth]{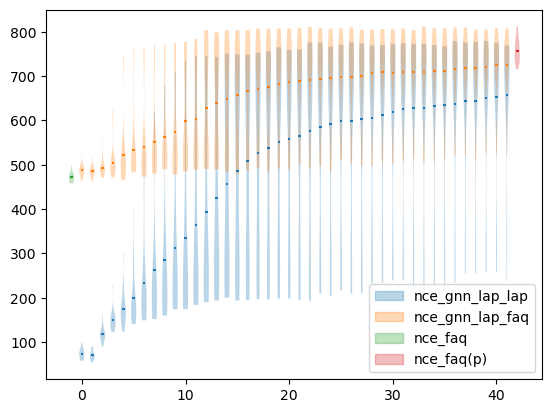}
  \caption{Sparse Erd\H{o}s-R\'enyi graphs: $\acc$ (top) and $\nce$ (bottom) as a function of the number of iterations $L$ at inference.}
  \label{fig:iter_sparse}
  \end{center}
\end{figure}

Figure \ref{fig:sparse} gives the performance of our chained GNNs trained at noise level $0.25$ for sparse Erd\H{o}s-R\'enyi graphs with average degree $d=4$ and size $n=500$. We observe that our chaining procedure is able to generalize to noise levels different from the one used during training and outperforms $\FAQ(D_\cx)$ in all cases.

\begin{table}[!t]
  \caption{Accuracy ($\acc$) defined in \eqref{eq:defmetrics} for sparse Erd\H{o}s-R\'enyi graphs as a function of the noise $p_\noise$. FGNN refers to the architecture in Section \ref{sec:architecture} and ChFGNN to our chained FGNNs. $\Proj$ and $\FAQ$ are used to produce a permutation (from the similarity matrix computed).}
  \label{tab:ER_sparse_acc}
\begin{center}
\begin{small}
\begin{sc}
  \begin{tabular}{lllllllllll}
    \toprule
   ER 4 ($\acc$) & noise & 0 & 0.05 & 0.1 & 0.15  & 0.2 & 0.25 & 0.3 & 0.35 \\
    \midrule
Baselines & $\Proj(D_\cx)$ & 98.0 & 97.3 & 90.3 & 59.3 & 23.3 & 9.1 & 4.0 & 2.0 \\
  & $\FAQ(D_\cx)$ & 98.0 & 97.5 & 96.3 & 94.6 & 72.9 & 13.0 & 3.7 & 1.7 \\
  \midrule
training 0.05 & ChFGNN $\Proj$ & 97.9 & 97.3 & 94.3 & 67.1 & 12.5 & 6.72 & 3.70 & 2.13 \\
  & ChFGNN $\FAQ$ & 97.9 & 97.5 & 96.2 & 72.9 & 43.5 & 10.8 & 4.04 & 1.69 \\
  \midrule
training 0.10 & ChFGNN $\Proj$ & 97.9 & 97.5 & 96.1 & 91.6 & 35.4 & 7.24 & 4.03 & 2.36 \\
  & ChFGNN $\FAQ$ & 97.9 & 97.5 & 96.4 & 94.0 & 38.6 & 12.3 & 4.73 & 1.91 \\
  \midrule
training 0.15 & ChFGNN $\Proj$ & 97.9 & 97.5 & 96.3 & 93.7 & 87.4 & 49.1 & 7.51 & 2.28 \\
  & ChFGNN $\FAQ$ & 97.9 & 97.5 & 96.4 & 94.3 & 90.3 & 54.7 & 9.71 & 1.79 \\
  \midrule
training 0.20 & ChFGNN $\Proj$ & 97.9 & 97.4 & 96.3 & 94.5 & 91.4 & 72.0 & 31.2 & 3.30 \\
  & ChFGNN $\FAQ$ & 97.9 & 97.5 & 96.4 & 95.2 & 93.1 & 76.3 & 35.0 & 3.39 \\
  \midrule
training 0.22 & ChFGNN $\Proj$ & 97.9 & 97.5 & 96.2 & 94.4 & 91.1 & 78.9 & 44.5 & 7.11 \\
  & ChFGNN $\FAQ$ & 97.9 & 97.5 & 96.4 & 95.3 & 93.1 & 82.1 & 48.3 & 7.78 \\
  \midrule
training 0.24 & ChFGNN $\Proj$ & 97.9 & 97.4 & 96.1 & 94.1 & 91.0 & 77.0 & 40.3 & 6.52 \\
  & ChFGNN $\FAQ$ & 97.9 & 97.5 & 96.4 & 95.3 & 93.3 & 80.1 & 43.3 & 6.93 \\
  \midrule
training 0.26 & ChFGNN $\Proj$ & 97.9 & 97.3 & 95.2 & 92.3 & 88.0 & 75.8 & 43.1 & 6.43 \\
  & ChFGNN $\FAQ$ & 97.9 & 97.5 & 96.4 & 95.2 & 93.2 & 82.5 & 48.2 & 6.87 \\
  \midrule
training 0.28 & ChFGNN $\Proj$ & 97.9 & 95.2 & 88.4 & 79.3 & 68.4 & 55.1 & 26.7 & 6.69 \\
  & ChFGNN $\FAQ$ & 97.9 & 97.5 & 96.3 & 94.9 & 92.7 & 82.9 & 38.7 & 7.62 \\
  \midrule
training 0.30 & ChFGNN $\Proj$ & 97.9 & 91.5 & 78.0 & 63.6 & 50.6 & 35.4 & 15.1 & 4.76 \\
  & ChFGNN $\FAQ$ & 97.9 & 97.4 & 96.2 & 94.8 & 92.1 & 73.5 & 23.4 & 5.00 \\
  \midrule
training 0.35 & ChFGNN $\Proj$ & 97.6 & 88.2 & 65.1 & 40.8 & 21.5 & 10.9 & 5.46 & 2.54 \\
  & ChFGNN $\FAQ$ & 97.9 & 97.4 & 96.2 & 94.5 & 68.0 & 19.4 & 5.79 & 2.00 \\
\bottomrule
  \end{tabular}
\end{sc}
\end{small}
\end{center}
\end{table}

\newpage
Each line in Tables \ref{tab:ER_sparse_acc} and \ref{tab:ER_sparse_nce} corresponds to a chained FGNN trained at a given level of noise (given on the left) and tested for all different noises.
\begin{table}[!t]
  \caption{Number of common edges ($\nce$) defined in \eqref{eq:defmetrics} for sparse Erd\H{o}s-R\'enyi graphs as a function of the noise $p_\noise$. FGNN refers to the architecture in Section \ref{sec:architecture} and ChFGNN to our chained FGNNs. $\Proj$ and $\FAQ$ are used to produce a permutation (from the similarity matrix computed).}
  \label{tab:ER_sparse_nce}
\begin{center}
\begin{small}
\begin{sc}
  \begin{tabular}{lllllllllll}
    \toprule
   ER 4 ($\nce$) & noise & 0 & 0.05 & 0.1 & 0.15  & 0.2 & 0.25 & 0.3 & 0.35 \\
    \midrule
    Baselines & $\Proj(D_\cx)$ & 997 & 950 & 853 & 499 & 195 & 130 & 115 & 112 \\
      & $\FAQ(D_\cx)$ & 997 & 950 & 898 & 847 & 723 & 504 & 487 & 485 \\
      \midrule
    training 0.05 & ChFGNN $\Proj$ & 997 & 950 & 885 & 630 & 116 & 95 & 87 & 83 \\
      & ChFGNN $\FAQ$ & 997 & 950 & 898 & 761 & 607 & 495 & 485 & 481 \\
      \midrule
    training 0.10 & ChFGNN $\Proj$ & 997 & 950 & 897 & 828 & 370 & 99 & 90 & 86 \\
      & ChFGNN $\FAQ$ & 997 & 950 & 899 & 845 & 606 & 501 & 487 & 483 \\
      \midrule
    training 0.15 & ChFGNN $\Proj$ & 996 & 950 & 898 & 840 & 768 & 511 & 254 & 86 \\
      & ChFGNN $\FAQ$ & 997 & 950 & 899 & 846 & 791 & 651 & 520 & 484 \\
      \midrule
    training 0.20 & ChFGNN $\Proj$ & 996 & 950 & 898 & 846 & 792 & 665 & 456 & 338 \\
      & ChFGNN $\FAQ$ & 997 & 950 & 899 & 849 & 800 & 715 & 596 & 529 \\
      \midrule
    training 0.22 & ChFGNN $\Proj$ & 997 & 950 & 898 & 845 & 790 & 694 & 503 & 319 \\
      & ChFGNN $\FAQ$ & 997 & 950 & 899 & 849 & 800 & 730 & 626 & 534 \\
      \midrule
    training 0.24 & ChFGNN $\Proj$ & 997 & 950 & 897 & 844 & 789 & 686 & 480 & 296 \\
      & ChFGNN $\FAQ$ & 997 & 950 & 899 & 849 & 800 & 726 & 613 & 527 \\
      \midrule
    training 0.26 & ChFGNN $\Proj$ & 997 & 949 & 892 & 834 & 770 & 672 & 499 & 338 \\
      & ChFGNN $\FAQ$ & 997 & 950 & 899 & 849 & 800 & 731 & 626 & 537 \\
      \midrule
    training 0.28 & ChFGNN $\Proj$ & 996 & 934 & 836 & 724 & 612 & 504 & 374 & 311 \\
      & ChFGNN $\FAQ$ & 997 & 950 & 899 & 848 & 799 & 732 & 599 & 530 \\
      \midrule
    training 0.30 & ChFGNN $\Proj$ & 996 & 897 & 726 & 566 & 446 & 345 & 271 & 246 \\
      & ChFGNN $\FAQ$ & 997 & 950 & 898 & 848 & 797 & 704 & 552 & 513 \\
      \midrule
    training 0.35 & ChFGNN $\Proj$ & 995 & 860 & 578 & 347 & 219 & 173 & 159 & 134 \\
      & ChFGNN $\FAQ$ & 997 & 950 & 898 & 847 & 702 & 524 & 494 & 489 \\
    \bottomrule
  \end{tabular}
\end{sc}
\end{small}
\end{center}
\end{table}

\newpage
\subsection{Additional results for dense Erd\H{o}s-R\'enyi graphs}

For the correlated dense Erd\H{o}s-R\'enyi graphs, we used the same dataset as in \cite{yu2023seedgnn} with $500$ nodes and an average degree of $80$. Again, with a noise level of $20\%$, our chaining GNNs clearly outperform the existing learning algorithms, as we obtain a perfect accuracy (as opposed to an accuracy of zero in \cite{yu2023seedgnn} and \cite{chen2020multi} without any seed). We see in Table \ref{tab:ER_dense_acc} that in this dense setting, $\FAQ(D_{\cx})$ is very competitive but is still slightly outperformed by our chaining FGNNs (orange curve with $\Proj$ and red curve with $\FAQ$, top). In terms of number of common edges, our chained FGNNs does not perform well with $\Proj$ but performs best with $\FAQ$, see Table \ref{tab:ER_dense_acc} where the level of noise used for training was $24\%$.

\begin{figure}[ht]
  \begin{center}
  \includegraphics[width=0.45\textwidth]{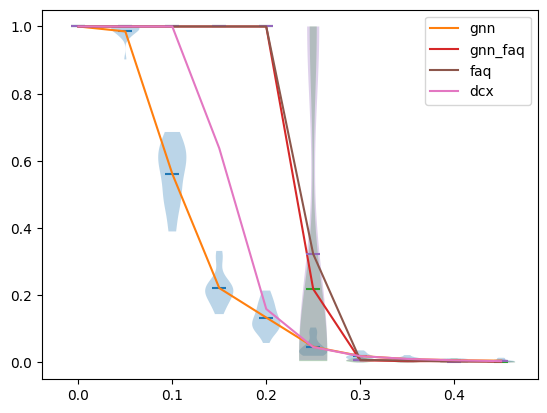}
  \includegraphics[width=0.46\textwidth]{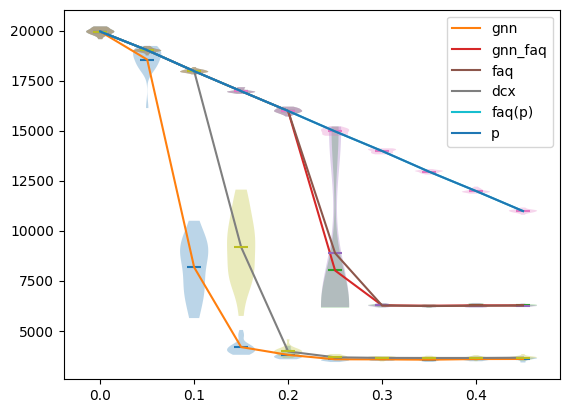}
  \caption{Dense Erd\H{o}s-R\'enyi graphs: $\acc$ (top) and $\nce$ (bottom) as a function of the noise level. Chained FGNNs were trained at noise level $0.24$. {\bf gnn} (resp.\ {\bf gnn\_faq}) for chained FGNNs with $\Proj$ (resp. $\FAQ$) as the last step. {\bf faq} for $\FAQ(D_\cx)$ and {\bf faq(p)} for $\FAQ(\pi^\star)$.}
  \label{fig:dense}
  \end{center}
\end{figure}

Figure \ref{fig:dense} gives the performance of our chained GNNs trained at noise level $0.24$ for dense Erd\H{o}s-R\'enyi graphs with average degree $d=80$ and size $n=500$.

\begin{table}[!t]
  \caption{Accuracy ($\acc$) defined in \eqref{eq:defmetrics} for dense Erd\H{o}s-R\'enyi graphs as a function of the noise $p_\noise$. FGNN refers to the architecture in Section \ref{sec:architecture} and ChFGNN to our chained FGNNs. $\Proj$ and $\FAQ$ are used to produce a permutation (from the similarity matrix computed).}
  \label{tab:ER_dense_acc}
\begin{center}
\begin{small}
\begin{sc}
  \begin{tabular}{lllllllllll}
    \toprule
   ER 80 ($\acc$) & noise & 0 & 0.05 & 0.1 & 0.15  & 0.2 & 0.25 & 0.3 & 0.35 \\
 \midrule
 Baselines & $\Proj(D_\cx)$ & 100. & 100. & 100. & 60.8 & 14.3 & 4.3 & 1.8 & 1.1 \\
   & $\FAQ(D_\cx)$ & 100. & 100. & 100. & 100. & 100. & 21.2 & 0.9 & 0.5 \\
   \midrule
 training 0.05 & ChFGNN $\Proj$ & 100. & 100. & 99.8 & 16.0 & 5.95 & 2.76 & 1.77 & 1.01 \\
   & ChFGNN $\FAQ$ & 100. & 100. & 100. & 100. & 54.4 & 1.16 & 0.72 & 0.47 \\
   \midrule
 training 0.10 & ChFGNN $\Proj$ & 100. & 100. & 100. & 88.9 & 7.49 & 3.47 & 1.99 & 1.16 \\
   & ChFGNN $\FAQ$ & 100. & 100. & 100. & 89.0 & 80.0 & 6.58 & 0.85 & 0.52 \\
   \midrule
 training 0.15 & ChFGNN $\Proj$ & 100. & 100. & 99.9 & 99.9 & 75.0 & 3.57 & 2.14 & 1.21 \\
   & ChFGNN $\FAQ$ & 100. & 100. & 100. & 100. & 75.1 & 3.85 & 0.85 & 0.55 \\
   \midrule
 training 0.20 & ChFGNN $\Proj$ & 100. & 100. & 100. & 99.9 & 94.0 & 22.9 & 2.06 & 1.22 \\
   & ChFGNN $\FAQ$ & 100. & 100. & 100. & 100. & 95.0 & 22.7 & 0.83 & 0.52 \\
   \midrule
 training 0.22 & ChFGNN $\Proj$ & 100. & 100. & 100. & 99.9 & 97.9 & 49.5 & 2.21 & 1.25 \\
   & ChFGNN $\FAQ$ & 100. & 100. & 100. & 100. & 99.0 & 50.5 & 1.00 & 0.52 \\
   \midrule
 training 0.24 & ChFGNN $\Proj$ & 100. & 99.9 & 93.5 & 83.4 & 67.4 & 34.0 & 2.16 & 1.33 \\
   & ChFGNN $\FAQ$ & 100. & 100. & 100. & 100. & 98.1 & 57.6 & 0.96 & 0.55 \\
   \midrule
 training 0.26 & ChFGNN $\Proj$ & 100. & 99.9 & 78.3 & 39.4 & 13.0 & 3.91 & 2.07 & 1.29 \\
   & ChFGNN $\FAQ$ & 100. & 100. & 100. & 100. & 94.1 & 5.75 & 0.82 & 0.54 \\
   \midrule
 training 0.28 & ChFGNN $\Proj$ & 100. & 99.8 & 70.7 & 31.2 & 9.88 & 3.94 & 2.01 & 1.19 \\
   & ChFGNN $\FAQ$ & 100. & 100. & 100. & 100. & 84.7 & 12.7 & 0.83 & 0.51 \\
   \midrule
 training 0.30 & ChFGNN $\Proj$ & 100. & 99.5 & 62.3 & 24.2 & 8.27 & 3.28 & 1.92 & 1.18 \\
   & ChFGNN $\FAQ$ & 100. & 100. & 100. & 100. & 80.7 & 3.24 & 0.81 & 0.49 \\
   \midrule
 training 0.35 & ChFGNN $\Proj$ & 100. & 96.1 & 47.8 & 18.3 & 6.86 & 3.47 & 1.92 & 1.14 \\
   & ChFGNN $\FAQ$ & 100. & 100. & 100. & 100. & 69.7 & 8.52 & 0.77 & 0.52 \\
 \bottomrule
  \end{tabular}
\end{sc}
\end{small}
\end{center}
\end{table}

\newpage
Each line in Tables \ref{tab:ER_dense_acc} and \ref{tab:ER_dense_nce} corresponds to a chained FGNN trained at a given level of noise (given on the left) and tested for all different noises.
\begin{table}[!t]
  \caption{Number of common edges ($\nce$) defined in \eqref{eq:defmetrics} for dense Erd\H{o}s-R\'enyi graphs as a function of the noise $p_\noise$. FGNN refers to the architecture in Section \ref{sec:architecture} and ChFGNN to our chained FGNNs. $\Proj$ and $\FAQ$ are used to produce a permutation (from the similarity matrix computed).}
  \label{tab:ER_dense_nce}
\begin{center}
\begin{small}
\begin{sc}
  \begin{tabular}{lllllllllll}
    \toprule
   ER 80 ($\nce$) & noise & 0 & 0.05 & 0.1 & 0.15  & 0.2 & 0.25 & 0.3 & 0.35 \\
    \midrule
    Baselines & $\Proj(D_\cx)$ & 19964 & 18987 & 17966 & 8700 & 3888 & 3646 & 3633 & 3624 \\
  & $\FAQ(D_\cx)$ & 19964 & 18987 & 17968 & 16990 & 15972 & 7922 & 6272 & 6276 \\
  \midrule
training 0.05 & ChFGNN $\Proj$ & 19964 & 18987 & 17941 & 3794 & 3457 & 3421 & 3429 & 3411 \\
  & ChFGNN $\FAQ$ & 19964 & 18987 & 17968 & 16990 & 11408 & 6244 & 6252 & 6253 \\
  \midrule
training 0.10 & ChFGNN $\Proj$ & 19964 & 18987 & 17968 & 15479 & 3522 & 3459 & 3453 & 3449 \\
  & ChFGNN $\FAQ$ & 19964 & 18987 & 17968 & 15811 & 13935 & 6681 & 6251 & 6257 \\
  \midrule
training 0.15 & ChFGNN $\Proj$ & 19964 & 18987 & 17967 & 16989 & 12842 & 3470 & 3469 & 3456 \\
  & ChFGNN $\FAQ$ & 19964 & 18987 & 17968 & 16990 & 13544 & 6421 & 6254 & 6256 \\
  \midrule
training 0.20 & ChFGNN $\Proj$ & 19964 & 18987 & 17968 & 16990 & 15216 & 6113 & 3483 & 3472 \\
  & ChFGNN $\FAQ$ & 19964 & 18987 & 17968 & 16990 & 15487 & 8172 & 6258 & 6254 \\
  \midrule
training 0.22 & ChFGNN $\Proj$ & 19964 & 18987 & 17968 & 16987 & 15701 & 9189 & 3644 & 3628 \\
  & ChFGNN $\FAQ$ & 19964 & 18987 & 17968 & 16990 & 15876 & 10614 & 6257 & 6263 \\
  \midrule
training 0.24 & ChFGNN $\Proj$ & 19964 & 18969 & 16241 & 13028 & 9561 & 6166 & 3615 & 3591 \\
  & ChFGNN $\FAQ$ & 19964 & 18987 & 17968 & 16990 & 15779 & 11227 & 6258 & 6255 \\
  \midrule
training 0.26 & ChFGNN $\Proj$ & 19964 & 18976 & 12528 & 5795 & 3925 & 3626 & 3591 & 3545 \\
  & ChFGNN $\FAQ$ & 19964 & 18987 & 17968 & 16990 & 15388 & 6515 & 6257 & 6257 \\
  \midrule
training 0.28 & ChFGNN $\Proj$ & 19964 & 18948 & 10846 & 4975 & 3756 & 3587 & 3542 & 3512 \\
  & ChFGNN $\FAQ$ & 19964 & 18987 & 17968 & 16990 & 14424 & 7207 & 6253 & 6256 \\
  \midrule
training 0.30 & ChFGNN $\Proj$ & 19964 & 18861 & 9289 & 4419 & 3651 & 3489 & 3478 & 3472 \\
  & ChFGNN $\FAQ$ & 19964 & 18987 & 17968 & 16990 & 14032 & 6354 & 6254 & 6258 \\
  \midrule
training 0.35 & ChFGNN $\Proj$ & 19964 & 17850 & 6943 & 4003 & 3578 & 3512 & 3492 & 3461 \\
  & ChFGNN $\FAQ$ & 19964 & 18987 & 17968 & 16990 & 12877 & 6853 & 6254 & 6256 \\
\bottomrule
  \end{tabular}
\end{sc}
\end{small}
\end{center}
\end{table}

\newpage
\subsection{Additional results for regular graphs}

\begin{figure}[!h]
  \begin{center}
  \includegraphics[width=0.45\textwidth]{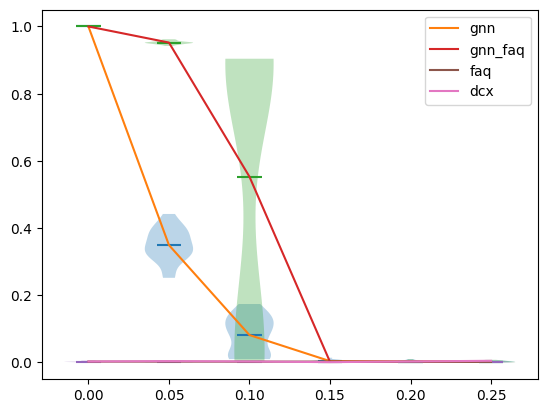}
  \includegraphics[width=0.46\textwidth]{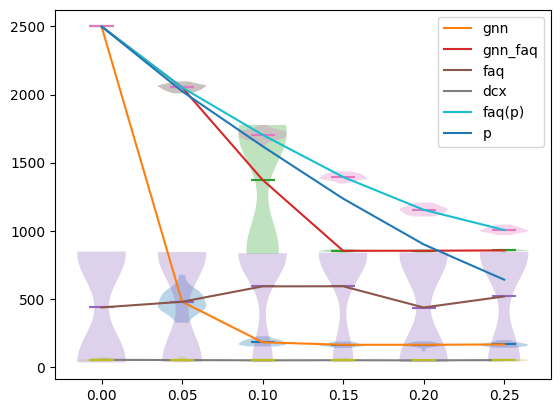}
  \caption{Regular graphs: $\acc$ (top) and $\nce$ (bottom) as a function of the noise level. Chained FGNNs were trained at noise level $0.1$. {\bf gnn} (resp.\ {\bf gnn\_faq}) for chained FGNNs with $\Proj$ (resp. $\FAQ$) as the last step. {\bf faq} for $\FAQ(D_\cx)$, {\bf faq(p)} for $\FAQ(\pi^\star)$, and {\bf p} for $\nce(\pi^\star)$.}
  \label{fig:reg}
  \end{center}
\end{figure}

Finally, we propose a new dataset of regular graphs with $500$ nodes and an average degree of $10$. This is a particularly challenging setting.
Indeed, Table \ref{tab:Reg_acc} shows that $\FAQ(D_{\cx})$ always fails to solve the graph matching problem here. The obstruction is not specific to regularity: it is the same obstruction captured by fractional isomorphism and by the 1-WL ceiling of MPNNs \cite{xu2018powerful}.
  \begin{theorem}\label{the:frac_iso}
    \cite{TINHOFER1991253} $G_A$ and $G_B$ are fractionally isomorphic, i.e. $\min_{D\in \Dcal_n}\|AD-DB\|^2_F =0$, if and only if 1-WL does not distinguish $G_A$ and $G_B$.
  \end{theorem}
\begin{proposition}[MPNN ceiling under fractional isomorphism]
\label{prop:mpnn-frac-iso}
Consider a Siamese MPNN encoder with shared parameters and identical initial node features on two fractionally isomorphic graphs $G_A$ and $G_B$. Then the final node embeddings are constant on the common stable 1-WL colour classes of the two graphs. Consequently any similarity matrix $S_{ij}=s(h_i^A,h_j^B)$ computed only from these embeddings is block-constant on pairs of 1-WL colour classes. A $\Proj$ decoder, or a $\FAQ$ initialization based only on $S$, can therefore use at most the common 1-WL partition and has no graph-dependent signal for distinguishing vertices inside a colour class.
\end{proposition}
\begin{proof}
Run 1-WL on both graphs from the same constant initial colour. By Theorem~\ref{the:frac_iso}, fractional isomorphism is equivalent to 1-WL indistinguishability, so the two graphs have the same stable colour histogram and the same common 1-WL colour classes up to relabelling of colours. The standard MPNN-versus-1-WL induction says that, for any fixed MPNN parameters, two vertices with the same 1-WL colour after $t$ refinement rounds have the same MPNN embedding after $t$ message-passing layers. Hence the final embeddings are constant on each stable colour class, with corresponding classes in $G_A$ and $G_B$ receiving the same embedding. Any pairwise score $s(h_i^A,h_j^B)$ therefore depends only on the colour classes of $i$ and $j$, not on their identities inside those classes. The assignment objective built from $S$ cannot distinguish permutations that only rearrange vertices within a common colour class.
\end{proof}
For regular graphs with constant features, the stable 1-WL partition has a single colour class, so Proposition~\ref{prop:mpnn-frac-iso} reduces to a constant similarity matrix: an unchained MPNN gives no more informative initialization than the barycenter $D_\cx=J$. The proposition does not rule out rank-augmented chained MPNNs, because the rank vector deliberately injects non-constant positional information after the first decoding step; empirically, however, our regular-graph experiments indicate that the 2-FWL FGNN backbone is the effective choice.
In contrast, our FGNN architecture defined in Section \ref{sec:architecture} is able to deal with regular graphs and our chaining procedure learns the correct information about the graph matching problem when the noise is low enough.

Note that we are in a setting where $\FAQ(\pi^\star)\neq \pi^*$ as soon as the noise level is above $5\%$ so that $\pi^\star\neq \pi^{A\to B}$. In this case, we believe that $\pi^{A\to B}=\FAQ(\pi^\star)$ (but to check it we should solve the graph matching problem!).
In Figure \ref{fig:reg}, the training was done with a noise level of $10\%$ so that labels were noisy. Still performances of our chained FGNNs with $\FAQ$ are very good. We do not know of any other algorithm working in this setting.

\begin{table}[!h]
  \caption{Accuracy ($\acc$) defined in \eqref{eq:defmetrics} for Regular graphs as a function of the noise $p_\noise$. FGNN refers to the architecture in Section \ref{sec:architecture} and ChFGNN to our chained FGNNs. $\Proj$ and $\FAQ$ are used to produce a permutation (from the similarity matrix computed).}
    \label{tab:Reg_acc}
    \begin{center}
    \begin{small}
    \begin{sc}
      Regular random graphs with degree $10$\\
  \begin{tabular}{llllllllll}
    \toprule
   Regular ($\acc$) & noise & 0 & 0.05 & 0.1 & 0.15  & 0.2 \\
    \midrule
Baselines & $\Proj(D_\cx)$ & 0.2 & 0.2 & 0.3 & 0.1 & 0.2 \\
  & $\FAQ(D_\cx)$ & 0.2 & 0.2 & 0.2 & 0.2 & 0.2 \\
  \midrule
training 0.05 & ChFGNN $\Proj$ & 100. & 95.2 & 2.60 & 0.67 & 0.27 \\
  & ChFGNN $\FAQ$ & 100. & 95.6 & 8.31 & 0.49 & 0.24 \\
  \midrule
training 0.07 & ChFGNN $\Proj$ & 100. & 95.3 & 34.6 & 0.70 & 0.27 \\
  & ChFGNN $\FAQ$ & 100. & 95.6 & 36.0 & 0.54 & 0.25 \\
  \midrule
training 0.09 & ChFGNN $\Proj$ & 100. & 95.2 & 54.4 & 0.86 & 0.34 \\
  & ChFGNN $\FAQ$ & 100. & 95.6 & 55.6 & 0.78 & 0.22 \\
  \midrule
training 0.11 & ChFGNN $\Proj$ & 100. & 72.4 & 30.5 & 0.86 & 0.27 \\
  & ChFGNN $\FAQ$ & 100. & 95.6 & 61.8 & 0.70 & 0.25 \\
  \midrule
training 0.13 & ChFGNN $\Proj$ & 79.2 & 16.9 & 2.13 & 0.55 & 0.25 \\
  & ChFGNN $\FAQ$ & 100. & 95.6 & 2.14 & 0.37 & 0.24 \\
  \midrule
training 0.15 & ChFGNN $\Proj$ & 60.4 & 13.3 & 1.69 & 0.52 & 0.30 \\
  & ChFGNN $\FAQ$ & 100. & 95.6 & 1.37 & 0.34 & 0.21 \\
\bottomrule
    \end{tabular}
  \end{sc}
\end{small}
\end{center}
\end{table}
\newpage
Each line in Tables \ref{tab:Reg_acc} and \ref{tab:Reg_nce} corresponds to a chained FGNN trained at a given level of noise (given on the left) and tested for all different noises.
\begin{table}[!h]
  \caption{Number of common edges ($\nce$) defined in \eqref{eq:defmetrics} for Regular graphs as a function of the noise $p_\noise$. FGNN refers to the architecture in Section \ref{sec:architecture} and ChFGNN to our chained FGNNs. $\Proj$ and $\FAQ$ are used to produce a permutation (from the similarity matrix computed).}
    \label{tab:Reg_nce}
    \begin{center}
    \begin{small}
    \begin{sc}
      Regular random graphs with degree $10$\\
  \begin{tabular}{llllllllll}
    \toprule
   Regular ($\nce$) & noise & 0 & 0.05 & 0.1 & 0.15  & 0.2 \\
    \midrule
    Baselines & $\Proj(D_\cx)$ & 51 & 51 & 50 & 49 & 50 \\
    & $\FAQ(D_\cx)$ & 385 & 425 & 456 & 369 & 496 \\
    \midrule
  training 0.05 & ChFGNN $\Proj$ & 2500 & 2034 & 178 & 101 & 100 \\
    & ChFGNN $\FAQ$ & 2500 & 2059 & 901 & 835 & 835 \\
    \midrule
  training 0.07 & ChFGNN $\Proj$ & 2500 & 2036 & 741 & 103 & 172 \\
    & ChFGNN $\FAQ$ & 2500 & 2059 & 1193 & 836 & 852 \\
    \midrule
  training 0.09 & ChFGNN $\Proj$ & 2500 & 2034 & 1105 & 281 & 95 \\
    & ChFGNN $\FAQ$ & 2500 & 2059 & 1381 & 871 & 836 \\
    \midrule
  training 0.11 & ChFGNN $\Proj$ & 2500 & 1343 & 563 & 192 & 114 \\
    & ChFGNN $\FAQ$ & 2500 & 2059 & 1438 & 850 & 837 \\
    \midrule
  training 0.13 & ChFGNN $\Proj$ & 1608 & 210 & 108 & 88 & 71 \\
    & ChFGNN $\FAQ$ & 2500 & 2059 & 841 & 836 & 834 \\
    \midrule
  training 0.15 & ChFGNN $\Proj$ & 984 & 163 & 96 & 86 & 87 \\
    & ChFGNN $\FAQ$ & 2500 & 2059 & 837 & 836 & 836 \\
  \bottomrule
    \end{tabular}
  \end{sc}
\end{small}
\end{center}
\end{table}

\subsection{LLM Usage}\label{sec:llmusage}
Large language models (LLMs) were employed in this work to assist with grammatical and syntactic corrections, to improve the clarity and readability of sentences and paragraphs, to support the generation of illustrative figures, to help refine one proof, and as coding assistants for the experimental pipeline and the ablation scripts. LLMs did not contribute to the core methodology or originality of the research.

\end{document}